%% file: main.tex
\documentclass[runningheads]{llncs}

\usepackage{eccv}

\usepackage{eccvabbrv}

\usepackage{graphicx}
\usepackage{booktabs}

\usepackage{import}
\subimport{./}{packages}
\subimport{./}{macros}

\usepackage{nicefrac}       

\usepackage[accsupp]{axessibility}  

\usepackage[breaklinks,colorlinks,citecolor=eccvblue,linkcolor=blue]{hyperref}

\newcommand{\Autoref}[1]{%
\begingroup%
\renewcommand{\equationautorefname}{Equation}%
\renewcommand{\figureautorefname}{Figure}%
\autoref{#1}%
\endgroup%
}

\renewcommand{\equationautorefname}{Eq.}
\renewcommand{\figureautorefname}{Fig.}

\usepackage{orcidlink}

\begin{document}

\title{Wavelet Convolutions for Large Receptive Fields} 

\author{Shahaf E. Finder\orcidlink{0000-0003-0254-1380} \and
Roy Amoyal\orcidlink{0009-0007-4524-0883} \and
Eran Treister\orcidlink{0000-0002-5351-0966} \and
Oren Freifeld\orcidlink{0000-0001-9816-9709}
}

\authorrunning{S.E.~Finder et al.}

\institute{The Department of Computer Science, Ben-Gurion University of the Negev, Israel\\
\email{\{finders,amoyalr\}@post.bgu.ac.il, \{erant,orenfr\}@cs.bgu.ac.il}}

\maketitle

\begin{abstract}
   \subimport{./}{abstract}
\end{abstract}
\subimport{./}{intro}
\subimport{./}{related}
\subimport{./}{method}
\subimport{./}{results}
\subimport{./}{conclusion}

\section*{Acknowledgements}
This work was supported in part by the Lynn and William Frankel Center at BGU CS, by Israel Science Foundation Personal Grant \#360/21, and by the Israeli Council for Higher Education (CHE) via the Data Science Research Center at BGU. 
S.E.F.'s work was also supported by the Kreitman School of Advanced Graduate Studies and by the BGU’s Hi-Tech Scholarship.

\bibliographystyle{splncs04}
\bibliography{refs}

\clearpage
\appendix
\subimport{./}{appendix}

\end{document}

%% file: packages.tex
\usepackage{graphicx}

\usepackage{xspace} 
\usepackage{amsmath,amssymb,amsfonts}
\usepackage{bbm}
\usepackage{bm}

\usepackage{subcaption}
\usepackage{enumitem}
\usepackage{color,soul}
\usepackage{xcolor}  

\usepackage{pifont}

\usepackage{tikz}
\usepackage{pgfplots}

\definecolor{lineBlue}{RGB}{57,106,177}
\definecolor{lineOrange}{RGB}{218,124,48}
\definecolor{lineGreen}{RGB}{62,150,81}
\definecolor{lineRed}{RGB}{204,37,41}
\definecolor{lineGray}{RGB}{83,81,84}
\definecolor{linePurple}{RGB}{107,76,154}
\definecolor{lineMaroon}{RGB}{146,36,40}
\definecolor{barBlue}{RGB}{114,147,203}
\definecolor{barOrange}{RGB}{225,151,76}
\definecolor{barGreen}{RGB}{132,186,91}
\definecolor{barRed}{RGB}{211,94,96}
\definecolor{barGray}{RGB}{128,133,133}
\definecolor{barPurple}{RGB}{144,103,167}
\definecolor{barMaroon}{RGB}{171,104,81}

\pgfplotsset{compat=1.18}
\usetikzlibrary{3d,decorations.text,shapes.arrows,positioning,fit,backgrounds}
\usetikzlibrary{intersections, pgfplots.fillbetween}

\usepackage{multirow}

\usetikzlibrary{arrows.meta}

\usepackage{algorithm}
\usepackage{algorithmic}

%% file: macros.tex
\makeatletter
\DeclareRobustCommand\onedot{\futurelet\@let@token\@onedot}
\def\@onedot{\ifx\@let@token.\else.\null\fi\xspace}

\def\eg{\emph{e.g}\onedot} \def\Eg{\emph{E.g}\onedot}
\def\ie{\emph{i.e}\onedot}

\def\wrt{w.r.t\onedot}

\makeatother


%




\definecolor{brown}{RGB}{165,42,42}

\bmdefine\ba{a}
\bmdefine\bb{b}
\bmdefine\bc{c}
\bmdefine\bd{d}
\bmdefine\be{e}
\bmdefine\boldf{f}
\bmdefine\bg{g}
\bmdefine\bh{h}
\bmdefine\bi{i}
\bmdefine\bj{j}
\bmdefine\bk{k}
\bmdefine\bl{l}
\bmdefine\bm{m}
\bmdefine\bn{n}
\bmdefine\bo{o}
\bmdefine\bp{p}
\bmdefine\bq{q}
\bmdefine\br{r}
\bmdefine\bs{s}
\bmdefine\bt{t}
\bmdefine\bu{u}
\bmdefine\bv{v}
\bmdefine\bw{w}
\bmdefine\bx{x}
\bmdefine\by{y}
\bmdefine\bz{z}
\bmdefine\bA{A}
\bmdefine\bB{B}
\bmdefine\bC{C}
\bmdefine\bD{D}
\bmdefine\bE{E}
\bmdefine\bF{F}
\bmdefine\bG{G}
\bmdefine\bH{H}
\bmdefine\bI{I}
\bmdefine\bJ{J}
\bmdefine\bK{K}
\bmdefine\bL{L}
\bmdefine\bM{M}
\bmdefine\bN{N}
\bmdefine\bO{O}
\bmdefine\bP{P}
\bmdefine\bQ{Q}
\bmdefine\bR{R}
\bmdefine\bS{S}
\bmdefine\bT{T}
\bmdefine\bU{U}
\bmdefine\bV{V}
\bmdefine\bW{W}
\bmdefine\bX{X}
\bmdefine\bY{Y}
\bmdefine\bZ{Z}

\bmdefine\balpha{\alpha}
\bmdefine\bbeta{\beta}
\bmdefine\bgamma{\gamma}

\bmdefine\bdelta{\delta}
\bmdefine\btheta{\theta}
\bmdefine\blambda{\lambda}
\bmdefine\bphi{\phi}
\bmdefine\bxi{\xi}
\bmdefine\bzeta{\zeta}
\bmdefine\boldeta{\eta}
\bmdefine\bpi{\pi}

\bmdefine\bmu{\mu}
\bmdefine\brho{\rho}
\bmdefine\bomega{\omega}
\bmdefine\bOmega{\Omega}
\bmdefine\bPi{\Pi}

\bmdefine\bvarepsilon{\varepsilon}
\bmdefine\bepsilon{\epsilon}

\bmdefine\bDelta{\Delta}

\bmdefine\bTheta{\Theta}

\bmdefine\bsigma{\sigma}
\bmdefine\bSigma{\Sigma}
\bmdefine\bPsi{\Psi}
\bmdefine\bLambda{\Lambda}

\bmdefine\bzero{0}
\bmdefine\bone{1}
\bmdefine\binfty{\infty}

\newcommand{\Ocal}{\mathcal{O}}

%% file: abstract.tex
In recent years, there have been attempts to increase the kernel size of Convolutional Neural Nets (CNNs) to mimic the global receptive field of Vision Transformers' (ViTs) self-attention blocks.
That approach, however, quickly hit an upper bound and saturated way before achieving a global receptive field. 
In this work, we demonstrate that by leveraging the Wavelet Transform (WT), it is, in fact, possible to obtain very large receptive fields without suffering from over-parameterization, \eg, for a $k \times k$ receptive field, the number of trainable parameters in the proposed method grows only logarithmically with $k$. 
The proposed layer, named WTConv, can be used as a drop-in replacement in existing architectures, results in an effective multi-frequency response, and scales gracefully with the size of the receptive field.
We demonstrate the effectiveness of the WTConv layer within ConvNeXt and MobileNetV2 architectures for image classification, as well as backbones for downstream tasks, and show it yields additional properties such as robustness to image corruption and an increased response to shapes over textures.
Our code is available at \url{https://github.com/BGU-CS-VIL/WTConv}.

%% file: intro.tex
\section{Introduction}
In the past decade, Convolutional Neural Networks (CNNs) have largely dominated many areas of computer vision. Nonetheless, with the recent emergence of Vision Transformers (ViTs)~\cite{Dosovitskiy:2020:vit}, which are an adaptation of the Transformer architecture~\cite{Vaswani:NIPS:2017:attention} used in natural language processing, CNNs have faced stiff competition. Concretely, 
the edge that ViTs are now believed to have over CNNs is primarily attributed to their multi-head self-attention layer.  This layer facilitates the global mixing of features, in contrast to convolutions that are, by construction, limited to local mixing of features. 
Consequently, several recent works have attempted to bridge the performance gap between CNNs and ViTs. Liu \etal~\cite{Liu:CVPR:2022:convnext} reconstructed the ResNet architecture and its training routine to keep up with the Swin Transformer~\cite{Liu:ICCV:2021:swintransformer}. One of the improvements in~\cite{Liu:CVPR:2022:convnext} was increasing the convolutions' kernel size. Empirically, however, that approach saturated at a kernel size of $7\times7$, meaning that increasing the kernel further did not help, and at some point even started deteriorating the performance. 
While naively increasing the size beyond $7\times7$ was not useful, Ding \etal~\cite{Ding:CVPR:2022:ScalingUpKernels} have shown that one can behoove from even larger kernels if those are better constructed.
Even then, however, eventually the kernels become over-parameterized and the performance saturates way before reaching a global receptive field. 

One intriguing property analyzed in~\cite{Ding:CVPR:2022:ScalingUpKernels} is that using larger kernels renders CNNs more shape-biased, which means that their ability to capture low frequencies in the image is improved. This finding was somewhat surprising because convolutional layers usually tend to respond to high-frequencies in the input~\cite{Wang:CVPR:2020:cnnshighfreq, Geirhos:ICLR:2018:CNNsTextureBiased, Tuli:2021:convvsattention, Gavrikov:CVPR:2024:biasesinimagenet}. 
This is unlike attention heads, which are known to be more attuned to low frequencies, as demonstrated in other studies~\cite{Park:ICLR:2022:HowViTWork, Naseer:NIPS:2021:propertiesofvit, Tuli:2021:convvsattention}. 

\subimport{./figures/}{erf}

The discussion above raises a natural question: 
Can we utilize signal-processing tools to increase the receptive field of convolutions effectively without suffering from over-parametrization?
In other words, can we have very large filters -- \eg, with a global receptive field -- and still improve the performance?
This paper offers a positive answer to this question. Our proposed approach leverages the Wavelet Transform  (WT)~\cite{Daubechies:SIAM:1992:tenlectures}, an established tool from time-frequency analysis, to make the receptive field of convolutions scale up well and, by cascading, also guide the CNNs to better respond to low frequencies.  
In part, our motivation to base our solution on the WT is that (unlike, \eg, the Fourier Transform) it retains some spatial resolution. This makes spatial operations (\eg, convolutions) in the wavelet domain more meaningful. 

More concretely, we propose WTConv, a layer that uses the cascade WT decomposition and performs a set of small-kernel convolutions, each focusing on different frequency bands of the input in an increasingly larger receptive field. 
This process allows us to put more emphasis on low frequencies in the input while adding only a small number of trainable parameters. 
In fact, for a $k\times k$ receptive field, our number of trainable parameters grows only logarithmically with $k$. This fact, which is in contrast to some recent methods whose corresponding growth is quadratic, lets us obtain effective CNNs with an unprecedented effective receptive field (ERF)~\cite{Luo:NIPS:2016:erf} size (see~\autoref{fig:erf}).  

We design WTConv as a drop-in replacement for depth-wise convolutions that can be used as is within any given CNN architecture without additional modifications. 
We validate the effectiveness of WTConv by incorporating it into ConvNeXt~\cite{Liu:CVPR:2022:convnext} for image classification, demonstrating its utility in a fundamental vision task. 
Further leveraging ConvNeXt as a backbone, we extend our evaluation to more complex applications: using it within UperNet~\cite{Xiao:ECCV:2018:upernet} for semantic segmentation and within Cascade Mask R-CNN~\cite{Cai:PAMI:2019:cascademaskrcnn} for object detection.
In addition, we analyze additional benefits WTConv provides to CNNs.

\textbf{To summarize, our key contributions are:}
\begin{itemize}[leftmargin=2em,itemsep=0pt,topsep=0pt]
    \item A new layer, called WTConv, that uses the WT to increase the receptive field of convolutions effectively.
    \item WTConv is designed to be a drop-in replacement (for depth-wise convolutions) within given CNNs.
    \item Extensive empirical evaluation demonstrates that WTConv improves CNNs' results in several key computer-vision tasks.
    \item Analysis of WTConv's contribution to CNN's scalability, robustness, shape-bias, and ERF.
\end{itemize}

%% file: figures/erf.tex
\begin{figure}[t]
    \centering
    \setlength\tabcolsep{10pt}
    \begin{tabular}{ccc}
    \begin{subfigure}[b]{0.245\textwidth}
        \includegraphics[width=\textwidth]{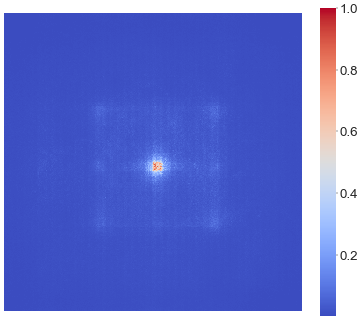}
        \caption{RepLK}
    \end{subfigure}&
    \begin{subfigure}[b]{0.245\textwidth}
        \includegraphics[width=\textwidth]{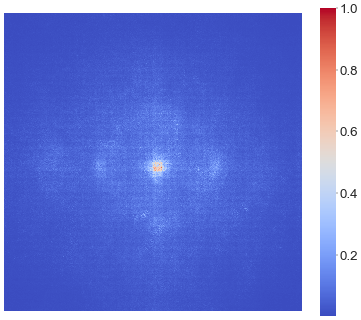}
        \caption{SLaK}
    \end{subfigure}&
    \begin{subfigure}[b]{0.245\textwidth}
        \includegraphics[width=\textwidth]{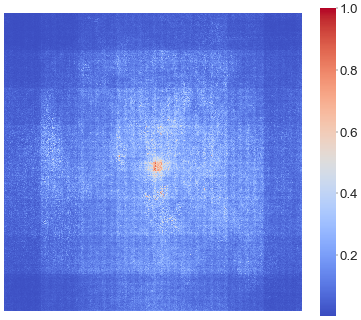}
        \caption{WTConv}
    \end{subfigure}
    \end{tabular}
    \caption{The Effective Receptive Fields \cite{Luo:NIPS:2016:erf} of ConvNeXt-T \cite{Liu:CVPR:2022:convnext} with different depth-wise convolutions.
    Evidently, the proposed WTConv achieves the largest field despite using fewer trainable parameters. This improves the convolution's ability to capture low frequencies and thus increases (\ie, improves) its shape bias, among other advantages.
    }
    \label{fig:erf}
\end{figure}

%% file: related.tex
\section{Related Work}\label{sec:relatedwork}
\subsection{Wavelet Transforms in Deep Learning}\label{subsec:WTinDL}
The WT~\cite{Daubechies:SIAM:1992:tenlectures}, a powerful tool for signal processing and analysis, has been widely used since the 1980s.
Following its success in classical settings, recently the WT has also been incorporated into neural network architectures for a variety of tasks.
Wang \etal~\cite{Wang:entropy:2021:ecgclasswavelet} extract features from the time-frequency components of ECG signals.
Huang \etal~\cite{Huang:ICCV:2017:waveletsuperres} and Guo \etal~\cite{Guo:CVPRw:2017:waveletforsr} predict wavelet high-frequency coefficients of an input image to reconstruct a higher-resolution output.
Duan \etal~\cite{Duan:PR:2017:sar} and Williams and Li~\cite{Williams:ICLR:2018:waveletpooling} use WTs as pooling operators within CNNs.
Gal \etal~\cite{Gal:TOG:2021:swagan}, Guth \etal~\cite{Guth:NIPS:2022:WaveletGenMod}, and Phung \etal~\cite{Phung:CVPR:2023:waveletdiffusion} use wavelets in generative models to enhance the visual quality of generated images as well as to improve computational performance.
Finder \etal~\cite{Finder:NIPS:2022:WCC} utilize wavelets to compress feature maps for more efficient CNNs. 
Saragadam \etal~\cite{Saragadam:CVPR:2023:wire} use wavelets as activation functions for implicit neural representations.

More related to our work, Liu \etal~\cite{Liu:CVPR:2018:multilevelwaveletcnn} and Alaba \etal~\cite{Alaba:2022:wcnn3d} utilize the WT in a modified U-Net architecture~\cite{Ronneberger:MICCAI:2015:unet} for down-sampling and the inverse WT for up-sampling. In another work related to ours, Fujieda \etal~\cite{Fujieda:2018:wcnn} propose a DenseNet-type architecture that uses wavelets to reintroduce lower frequencies in the input to later layers.
Although not wavelet-related, Chen \etal~\cite{Chen:ICCV:2019:octave} propose performing convolutions on multi-resolution input by initially separating the image to high and low resolutions and performing information exchange between the two along the network.
These works exemplify the benefits of performing convolutions on the input's low-frequency components separately from the high-frequency components for a more informative feature map. This feature motivates our work as well. However, the methods from \cite{Liu:CVPR:2018:multilevelwaveletcnn,Fujieda:2018:wcnn,Alaba:2022:wcnn3d} are highly customized architectures and cannot seamlessly be used within other CNN architectures, while \cite{Chen:ICCV:2019:octave} focuses on computational efficiency.
In contrast, we propose a lighter, easier-to-user, and linear layer that can be used as a drop-in replacement for depth-wise convolutions and that results in an improved receptive field. Importantly, our method can fit within any network that uses depth-wise convolution and, therefore, is not limited to a single task.

\subsection{Large-Kernel Convolutions} \label{subsec:lkconv}
When regarding convolution configurations, VGG~\cite{Simonyan:CoRR:2015:VGG} has set the standard for modern CNNs by using $3\times3$ convolutions, sacrificing the size of single-layer receptive field to increase the network's depth (from under 10 layers to around 20). Since then, with increased computations and improved architectures, CNNs have become much deeper, but the kernel-size parameter was left largely unexplored.

One major change to the traditional convolutions was the introduction of separable convolutions~\cite{Wang:CVPR:2017:factorizedconvolution, Vanhoucke:ICLR:2014:learningvisrep}. 
The separable convolutions were popularized by Xception~\cite{Chollet:CVPR:2017:xception} and MobileNet~\cite{Howard:2017:mobilenets} and are adopted in most modern architectures~\cite{Liu:CVPR:2022:convnext, Sandler:CVPR:2018:mobilenetv2}. In this approach, spatial convolutions are performed per channel (\ie, depth-wise), and the cross-channel operations are performed using $1\times1$ kernels (\ie, point-wise). 
This separation of convolutions also creates a degree of separation between the kernel size and the channel dimension \wrt the number of parameters and operations.
Each spatial convolution of kernel size $k$ and $c$ channels now has only $k^2 \cdot c$ parameters (instead of $k^2 \cdot c^2$), and this allows it to scale better with $k$, albeit still quadratically.

Concurrently, the introduction to vision tasks~\cite{Dosovitskiy:2020:vit, Liu:ICCV:2021:swintransformer} of transformers -- with their non-local self-attention layers -- has typically yielded better results than those of the local-mixing convolutions. 
This, together with the aforementioned recent use of separable convolutions, has reignited the interest in exploring larger kernels for CNNs. 
In particular, Liu \etal~\cite{Liu:CVPR:2022:convnext} have reexamined the popular ResNet architecture~\cite{He:CVPR:2016:resnet}, including an empirical comparison of different kernel sizes, concluding that the performance  saturated at a kernel size of $7\times7$. Trockman and Kolter~\cite{Trockman:2022:patches} have attempted to mimic the ViT architecture using only convolutions and have shown impressive results by using $9\times9$ convolutions to replace the attention (or ``mixer") component.
Ding \etal~\cite{Ding:CVPR:2022:ScalingUpKernels} have suggested that simply increasing the size of the kernel hurts the locality property of convolutions. Thus, they proposed using a small kernel in parallel to a large one and then summing their outputs. With that technique, they successfully trained CNNs with kernel sizes up to $31\times31$. Liu \etal~\cite{Liu:ICLR:2023:slak} have successfully increased their kernels to size $51\times51$ by factorizing it to a set of parallel $51\times5$ and $5\times51$ kernels. In addition, they introduced sparsity together with the expansion of the width of the network. 
However, this idea of using more channels (with sparsity) is orthogonal to increasing the kernel size.
While our work partly draws inspiration from~\cite{Ding:CVPR:2022:ScalingUpKernels, Liu:ICLR:2023:slak}, in our case the proposed layer sums outputs from various frequency components of the input, capturing multiple receptive fields. 

Another way to achieve a global receptive field is performing the spatial mixing in the frequency domain following a Fourier transform (\eg~\cite{Ra:NIPS2021:gfnet, chi:nips:2020:ffc, Haber:ICML:2019:IMEX}). However, the Fourier transform converts the input to be represented entirely in the frequency domain, therefore it fails to learn local interactions between neighboring pixels. In contrast, the WT successfully preserves some local information while decomposing the image to different frequency bands, allowing us to operate on different levels of decomposition. Moreover, Fourier-based methods tend to rely on a specific size of input for the number of weights and, therefore, can be difficult to use for downstream tasks.
A concurrent work \cite{Grabinski:TMLR:2024:NIFF} utilizes neural implicit functions for efficient mixing in the frequency domain.

%% file: method.tex
\section{Method}
In this section, we first describe how the wavelet transform is performed using convolutions, and then we propose our solution for performing convolution in the wavelet domain, named WTConv. We also describe WTConv's theoretical benefits and analyze its computational cost.

\subsection{Preliminaries: The Wavelet Transform as Convolutions}\label{subsec:wt_as_conv}

In this work, we employ the Haar WT as it is efficient and straightforward \cite{Finder:NIPS:2022:WCC,Gal:TOG:2021:swagan,Huang:ICCV:2017:waveletsuperres}. However, we note that our approach is not limited to it as other wavelet bases can be used, albeit at an increased computational cost.

Given an image $X$, the one-level Haar WT over one spatial dimension (width or height) is given by a depth-wise convolution with the kernels $[1,1]/\sqrt{2}$ and $[1,-1]/\sqrt{2}$ followed by a standard downsampling operator of factor $2$. To perform the 2D Haar WT, we  compose the operation on both dimensions, resulting in a depth-wise convolution with a stride of 2 using the following set of four filters:
\begin{align}\label{eq:HaarKernel}
\begin{split}
    f_{LL} = \frac{1}{2}
    \begin{bmatrix}
    1 & 1 \\
    1 & 1
    \end{bmatrix},\,
    f_{LH} = \frac{1}{2}
    \begin{bmatrix}
    1 & -1 \\
    1 & -1
    \end{bmatrix},\, 
    f_{HL} = \frac{1}{2}
    \begin{bmatrix}
    \;\;1 & \;\;1 \\
    -1 & -1
    \end{bmatrix},\,
    f_{HH} = \frac{1}{2}
    \begin{bmatrix}
    \;\;1 & -1 \\
    -1 & \;\;1
    \end{bmatrix}.
\end{split}
\end{align}
Note that $f_{LL}$ is a low-pass filter, and $f_{LH},f_{HL},f_{HH}$ are a set of high-pass filters. For each input channel, the output of the convolution 
\begin{align}
\begin{split}
\left[X_{LL},X_{LH},X_{HL},X_{HH}\right] = \mbox{Conv}([f_{LL},f_{LH},&f_{HL},f_{HH}],X)    
\end{split}
\end{align}
has four channels, each of which has (in each spatial dimension) half the resolution of $X$.
$X_{LL}$ is the low-frequency component of $X$, while $X_{LH}, X_{HL}, X_{HH}$ are its horizontal, vertical, and diagonal high-frequency components.

Since the kernels in \autoref{eq:HaarKernel} form an orthonormal basis, applying the inverse wavelet transform (IWT) is obtained by the transposed convolution: 
\begin{align}
\begin{split}
X = \mbox{Conv-transposed}(&\left[f_{LL},f_{LH},f_{HL},f_{HH}\right],\\
&\left[X_{LL},X_{LH},X_{HL},X_{HH}\right]).     
\end{split}
\end{align}

The cascade wavelet decomposition is then given by recursively decomposing the low-frequency component. Each level of the decomposition is given by
\begin{align}\label{eq:cascadewt}
    X^{(i)}_{LL}, X^{(i)}_{LH}, X^{(i)}_{HL}, X^{(i)}_{HH} = \mathrm{WT}(X^{(i-1)}_{LL}) 
\end{align}
where $X^{(0)}_{LL} = X$ and $i$ is the current level. This results in an increased frequency resolution and a reduced spatial resolution for the lower frequencies.

\subimport{./figures/}{conv_in_wavelet_domain}

\subsection{Convolution in the Wavelet Domain} \label{subsec:convinwt}
As described in \autoref{subsec:lkconv}, increasing the kernel size of a convolutional layer increases the number of parameters (and, therefore, the degrees of freedom) quadratically. %
To mitigate that, we propose the following. First, use the WT to filter and downscale the lower- and higher-frequency content of the input. Then, perform a small-kernel depth-wise convolution on the different frequency maps before using the IWT to construct the output.
In other words, the process is given by
\begin{align}
    Y = \mathrm{IWT}(\mathrm{Conv}(W,\mathrm{WT}(X))),
\end{align}
where $X$ is the input tensor, and $W$ is the weight tensor of a $k \times k$ depth-wise kernel with four times as many input channels as $X$.  
This operation not only separates the convolution between the frequency components but also allows a smaller kernel to operate in a larger area of the original input, \ie, increasing its receptive field \wrt the input. See \autoref{fig:convinwavelet} for an illustration.

We take this 1-level combined operation and increase it further by using the same cascade principle from \autoref{eq:cascadewt}. The process is given by:
\begin{align}
    X^{(i)}_{LL},X^{(i)}_{H} &= \mathrm{WT}(X^{(i-1)}_{LL}),\\
    Y^{(i)}_{LL},Y^{(i)}_{H} &= \mathrm{Conv}(W^{(i)},(X^{(i)}_{LL},X^{(i)}_{H})),
\end{align}
where $X^{(0)}_{LL}$ is the input of the layer, and $X^{(i)}_H$ represents all three high-frequency maps of level $i$ described in~\autoref{subsec:wt_as_conv}.

To combine the outputs of the different frequencies, we use the fact that the WT and its inverse are linear operations, meaning that $\mathrm{IWT}(X+Y) = \mathrm{IWT}(X)+\mathrm{IWT}(Y)$. Therefore, performing
\begin{align}
    Z^{(i)} &= \mathrm{IWT}(Y^{(i)}_{LL}+Z^{(i+1)},Y^{(i)}_{H})
\end{align}
results in the summation of the different levels' convolutions, where $Z^{(i)}$ is the aggregated outputs from level $i$ onward. This is in line with \cite{Ding:CVPR:2022:ScalingUpKernels}, where two outputs of different-sized convolutions are summed as the output.

\subimport{./figures/}{example_images}

In contrast to \cite{Ding:CVPR:2022:ScalingUpKernels}, we can not normalize each of the $Y^{(i)}_{LL}, Y^{(i)}_H$, as the separate normalization of those does not correspond to normalization in the original domain. Instead, we find it is sufficient to perform only a channel-wise scaling to weigh each frequency component's contribution.
\Autoref{fig:wtconv_arch} visualizes WTConv for the case of a 2-level WT. The algorithm is provided in \textbf{Appendix A}.

\subsection{The Benefits of Using WTConv}

There are two main technical benefits of incorporating WTConv within a given CNN.
First, each level of WT increases the size of the layer's receptive field with only a small increase in the number of trainable parameters.
That is, the $\ell$-level cascading frequency decomposition of the WT, together with a fixed-size kernel, $k$, for each level, allows the number of parameters to scale linearly in the number of levels ($\ell\cdot4\cdot c\cdot k^2$) while the receptive field grows exponentially ($2^\ell\cdot k$).

The second benefit is that the WTConv layer is constructed to capture low frequencies better than a standard convolution.
This is because the repeated WT decomposition of the low frequencies of the input emphasizes them and increases the layer's corresponding response. 
This discussion complements the analysis that convolutional layers are known to respond to high frequencies in the input \cite{Park:ICLR:2022:HowViTWork, Geirhos:ICLR:2018:CNNsTextureBiased}. By leveraging compact kernels on the multifrequency inputs, the WTConv layer places the additional parameters where they are most needed.

\emph{In addition to yielding improved results on standard benchmarks, these technical benefits translate to improvement in the network's scalability in comparison to large-kernel methods, robustness \wrt corruption and distribution shift, and responding more to shapes over textures.} We empirically confirm these assumptions in \autoref{subsec:analysis}.

\subsection{Computational Cost}

The computational cost, in terms of floating-point operations (FLOPs), of a depth-wise convolution is 
\begin{align}
    C\cdot K_W \cdot K_H \cdot N_W \cdot N_H \cdot \frac{1}{S_W} \cdot \frac{1}{S_H},
\end{align}
where $C$ is the number of input channels, $(N_W,N_H)$ is the spatial dimension of the input, $(K_W,K_H)$ is the kernel size, and $(S_W,S_H)$ is the stride in each dimension. 
For example, consider a single-channel input of spatial dimensions $512\times512$. Performing convolution with a kernel of size $7\times7$ results in $12.8M$ FLOPs, while a $31\times31$ kernel size results in $252M$ FLOPs.
Considering the WTConv set of convolutions, each wavelet-domain convolution is performed over a reduced spatial dimension in a factor of 2, albeit at four times as many channels as the original input. This results in a FLOP count of
\begin{align}\label{eq:wt_conv_flops}
    C \cdot K_W \cdot K_H \cdot \left(N_W \cdot N_H + \sum\limits_{i=1}^\ell 4\cdot\frac{N_W}{2^i} \cdot \frac{N_H}{2^i}\right),
\end{align}
where $\ell$ is the number of WT levels. Continuing the earlier example of input size $512\times512$, performing the multi-frequency convolutions of a 3-level WTConv with kernel size of $5\times5$ (which covers a receptive field of $40\times40=(5\cdot 2^3) \times (5\cdot 2^3) $) results in $15.1M$ FLOPs. Of course, one also needs to add the cost of the WT calculation itself. We note that when the Haar basis is used, WTs can be implemented in a highly efficient way~\cite{Finder:NIPS:2022:WCC}. That said, with a naive implementation using the standard convolution operation, the FLOP count of the WTs is 
\begin{align}
4C\cdot \sum\nolimits_{i=0}^{\ell-1} \frac{N_W}{2^i} \cdot \frac{N_H}{2^i},
\end{align}
since the four kernels are of size $2\times2$ with a stride of 2 on each spatial dimension and operate on each of the input channels (see~\autoref{subsec:wt_as_conv}). Likewise, a similar analysis shows that the IWT has the same FLOP count as the WT. Continuing the example, this results in an addition of $2.8M$ FLOPs for 3 levels of WT and IWT, totaling $17.9M$ FLOPs, which still represent a substantial saving over the standard depth-wise convolutions of similar receptive fields. 

%% file: figures/conv_in_wavelet_domain.tex
\begin{figure}[t]

\centering
\begin{tikzpicture}[x={(0.7em,0)},y={(0,-0.7em)}, font=\sffamily\scriptsize]

\draw[step=0.35em,barGray!70!white,ultra thin] (-1.5,0) grid (10.5,12);
\draw[lineGray, ultra thin] (-1.5,0) rectangle (10.5,12);
\node at (4.5,13) {$X$};

\draw[draw=lineBlue, fill=barBlue, fill opacity=0.6] (0.5,2) rectangle (6.5,8);

\draw[-Stealth](11,6) -- (13,6) node [pos=0.5, above, font=\scriptsize] {WT};

\draw[step=0.35em,barGray!70!white,ultra thin] (13.5,0) grid (19.5,6);
\draw[lineGray, ultra thin] (13.5,0) rectangle (19.5,6);
\node at (16.5,7) {$X^{(1)}_{LL}$};
\draw[step=0.35em,barGray!75!white,ultra thin] (13.5,8) grid (19.5,14);
\draw[lineGray, ultra thin] (13.5,8) rectangle (19.5,14);
\node at (16.5,15) {$X^{(1)}_{HL}$};
\draw[step=0.35em,barGray!70!white,ultra thin] (21,0) grid (27,6);
\draw[lineGray, ultra thin] (21,0) rectangle (27,6);
\node at (24,7) {$X^{(1)}_{LH}$};
\draw[step=0.35em,barGray!70!white,ultra thin] (21,8) grid (27,14);
\draw[lineGray, ultra thin] (21,8) rectangle (27,14);
\node at (24,15) {$X^{(1)}_{HH}$};

\draw[draw=lineBlue, fill=barBlue, fill opacity=0.6] (14.5,1) rectangle (17.5,4);

\draw[-Stealth](27.5,6) -- (29.5,6) node [pos=0.5, above, font=\scriptsize] {WT};

\draw[step=0.35em,barGray!70!white,ultra thin] (30,0) grid (33,3);
\draw[lineGray, ultra thin] (30,0) rectangle (33,3);
\node at (31.5,4) {$X^{(2)}_{LL}$};
\draw[step=0.35em,barGray!70!white,ultra thin] (30,5) grid (33,8);
\draw[lineGray, ultra thin] (30,5) rectangle (33,8);
\node at (31.5,9) {$X^{(2)}_{HL}$};
\draw[step=0.35em,barGray!70!white,ultra thin] (35,0) grid (38,3);
\draw[lineGray, ultra thin] (35,0) rectangle (38,3);
\node at (36.5,4) {$X^{(2)}_{LH}$};
\draw[step=0.35em,barGray!70!white,ultra thin] (35,5) grid (38,8);
\draw[lineGray, ultra thin] (35,5) rectangle (38,8);
\node at (36.5,9) {$X^{(2)}_{HH}$};

\draw[draw=lineBlue, draw opacity=0.8, fill=barBlue, fill opacity=0.6] (30.5,0.5) rectangle (32,2);

\draw[-stealth, lineBlue, thick] (31.25,0.5) to [out=160, in=20] (16,1);
\draw[-stealth, lineBlue, thick] (16,1) to [out=160, in=20] (3.5,2);

\end{tikzpicture}

\caption{Performing convolution in the wavelet domain results in a larger receptive field. In this example, a $3\times3$ convolution is performed on the low-frequency band of the second-level wavelet domain $X^{(2)}_{LL}$, resulting in a 9-parameter convolution that responds to lower frequencies of a $12\times12$ receptive field in the input $X$.}
\label{fig:convinwavelet}

\end{figure}
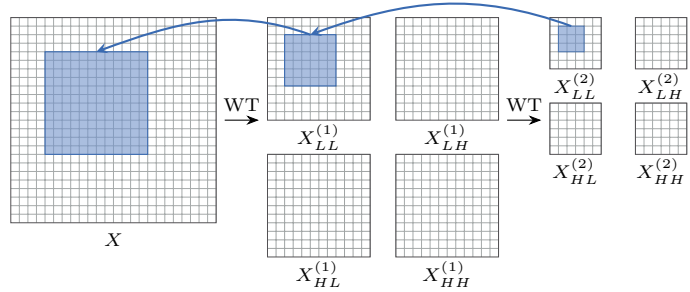

%% file: figures/example_images.tex
\begin{figure}[t]

    \tikzset{
      basic/.style  = {draw, text width=1cm, font=\sffamily, rectangle, align=center},
      wt/.style  = {basic, text width=1cm}
    }
    
    \centering
    \begin{tikzpicture}[x={(0.8cm,0)},y={(0,-0.8cm)}, font=\sffamily\small]
            
    \def\hdelta{0.1}
    \def\vdelta{0.1}
    
    \draw[-Stealth] (-1, 1.5) -- (0 - \hdelta, 1.5) node [pos=0.5, above, font=\footnotesize] {input};
    \node[inner sep=0pt] at (0,0) [anchor=north west]
        {\includegraphics[width=2.4cm]{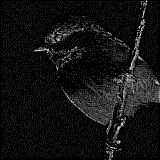}};
    
    \draw[-Stealth] (1.5, 3+\vdelta) -- (1.5, 3.75-\vdelta) node [pos=0.5, left, font=\footnotesize] {WT};
    \node[inner sep=0pt] at (0, 3.75) [anchor=north west]
        {\includegraphics[width=1.16cm]{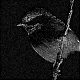}};
    \node[inner sep=0pt] at (0, 6.75) [anchor=south west]
        {\includegraphics[width=1.16cm]{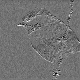}};
    \node[inner sep=0pt] at (3, 3.75) [anchor=north east]
        {\includegraphics[width=1.16cm]{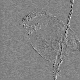}};
    \node[inner sep=0pt] at (3, 6.75) [anchor=south east]
        {\includegraphics[width=1.16cm]{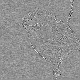}};
    
    \draw[-Stealth] (1.5, 6.75+\vdelta) -- (1.5, 7.05) -- (0.75, 7.05) -- (0.75, 7.5-\vdelta) node [pos=0.5, left, font=\footnotesize] {WT};
    \node[inner sep=0pt] at (0, 7.5) [anchor=north west]
        {\includegraphics[width=0.536cm]{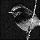}};
    \node[inner sep=0pt] at (0, 8.95) [anchor=south west]
        {\includegraphics[width=0.536cm]{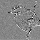}};
    \node[inner sep=0pt] at (1.45, 7.5) [anchor=north east]
        {\includegraphics[width=0.536cm]{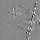}};
    \node[inner sep=0pt] at (1.45, 8.95) [anchor=south east]
        {\includegraphics[width=0.536cm]{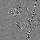}};
    
    \draw[-Stealth] (1.5+\hdelta, 8.225) -- (2.5-\hdelta, 8.225) node [pos=0.5, above, font=\footnotesize] {Conv};
    \node[inner sep=0pt] at (2.5, 7.5) [anchor=north west]
        {\includegraphics[width=0.536cm]{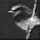}};
    \node[inner sep=0pt] at (2.5, 8.95) [anchor=south west]
        {\includegraphics[width=0.536cm]{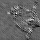}};
    \node[inner sep=0pt] at (3.95, 7.5) [anchor=north east]
        {\includegraphics[width=0.536cm]{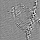}};
    \node[inner sep=0pt] at (3.95, 8.95) [anchor=south east]
        {\includegraphics[width=0.536cm]{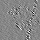}};
    
    \draw[-Stealth] (4+\hdelta, 8.225) -- (5-\hdelta, 8.225) node [pos=0.5, above, font=\footnotesize] {IWT};
    \node[inner sep=0pt] at (5, 7.5) [anchor=north west]
        {\includegraphics[width=1.16cm]{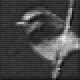}};
    
    \draw[-Stealth] (5.75, 7.5-\vdelta) -- (5.75, 7.05) -- (3.75, 7.05) -- (3.75, 5.25) node [pos=0.9, right, font=\footnotesize] {$+$};
    \draw[-Stealth] (3+\hdelta, 5.25) -- (4.5-\hdelta, 5.25) node [pos=0.5, above, font=\footnotesize] {Conv};
    \node[inner sep=0pt] at (4.5, 3.75) [anchor=north west]
        {\includegraphics[width=1.16cm]{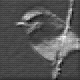}};
    \node[inner sep=0pt] at (4.5, 6.75) [anchor=south west]
        {\includegraphics[width=1.16cm]{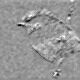}};
    \node[inner sep=0pt] at (7.5, 3.75) [anchor=north east]
        {\includegraphics[width=1.16cm]{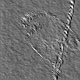}};
    \node[inner sep=0pt] at (7.5, 6.75) [anchor=south east]
        {\includegraphics[width=1.16cm]{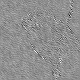}};
    
    \draw[-Stealth] (7.5+\hdelta, 5.25) -- (9-\hdelta, 5.25) node [pos=0.5, above, font=\footnotesize] {IWT};
    \node[inner sep=0pt] at (9, 3.75) [anchor=north west]
        {\includegraphics[width=2.4cm]{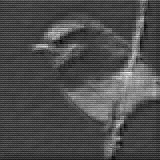}};
    
    \draw[-Stealth] (3+\hdelta, 1.5) -- (4.5-\hdelta, 1.5) node [pos=0.5, above, font=\footnotesize] {Conv};
    \node[inner sep=0pt] at (4.5, 0) [anchor=north west]
        {\includegraphics[width=2.4cm]{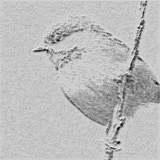}};
    
    \draw[-Stealth] (10.5, 3.75-\vdelta) -- (10.5, 3.35) -- (8.25, 3.35) -- (8.25, 1.5) node [pos=0.9, right, font=\footnotesize] {$+$};
    \draw[-Stealth] (7.5+\hdelta, 1.5) -- (9-\hdelta, 1.5);
    \node[inner sep=0pt] at (9, 0) [anchor=north west]
        {\includegraphics[width=2.4cm]{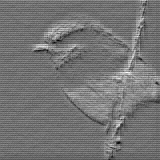}};
        
    \draw[-Stealth] (12 + \hdelta, 1.5) -- (13.2, 1.5) node [pos=0.5, above, font=\footnotesize] {output};
    
    \end{tikzpicture}
    
    \caption{An example of the WTConv operation on a single channel taken from the third inverted residual block of MobileNetV2 (see \autoref{subsec:analysis}) using a 2-level wavelet decomposition and $3\times3$ kernel sizes for the convolutions.}
    \label{fig:wtconv_arch}

\end{figure}

%% file: results.tex
\section{Results}\label{sec:results}
In this section, we experiment with WTConv in several settings. First, in \autoref{subsec:imagenet1k}, we train and evaluate ConvNeXt~\cite{Liu:CVPR:2022:convnext} with WTConv for ImageNet-1K~\cite{Deng:CVPR:2009:imagenet} classification. Then, in \autoref{subsec:semseg} and \autoref{subsec:det}, we use the trained models as backbones for downstream tasks. Finally, in \autoref{subsec:analysis}, we further analyze the contribution of WTConv to the network. 

\subsection{ImageNet-1K Classification}\label{subsec:imagenet1k}
For ImageNet-1K~\cite{Deng:CVPR:2009:imagenet}, we use ConvNeXt~\cite{Liu:CVPR:2022:convnext} as our base architecture and replace the $7\times7$ depth-wise convolution with WTConv.
ConvNeXt, as an extension of ResNet, mainly consists of four stages with downsampling operations between them. 
We set WTConv's levels to [5,4,3,2] and kernel size to $5\times5$ for these stages to achieve a global receptive field at each step for an input size of $224\times224$.
We use two training schedules of 120 and 300 epochs (see~\textbf{Appendix B} for details).

\subimport{./tables/}{imagenet_convnext_120}
\autoref{tab:convnextimagenet_short} shows the results for the 120-epoch schedule.
Since all networks use the same base ConvNeXt-T architecture, we report the number of parameters for the depth-wise (marked D-W) convolutions. 
Note that, for a fair comparison, we report SLaK's and VAN's results using only the depth-wise kernel decomposition, as we compare only the effect of increasing the receptive field.
We emphasize that WTConv achieves the best results while being the most parameter-efficient among the top-scoring methods. 
Moreover, it achieves a global receptive field with less than half of GFNet's number of parameters.

\subimport{./tables/}{imagenet_convnext_300}
In \autoref{tab:convnextimagenet_long}, which shows the results for the 300-epoch schedule, we compare WTConvNeXt to Swin~\cite{Liu:ICCV:2021:swintransformer} and ConvNeXt~\cite{Liu:CVPR:2022:convnext}.
\autoref{tab:convnextimagenet_short} and \autoref{tab:convnextimagenet_long} both show that introducing WTConv to ConvNeXt results in a substantial improvement in classification accuracy while introducing only a slight increase in parameters and FLOPs. \Eg, moving from ConvNeXt-S to ConvNeXt-B adds 39M parameters and 6.7 GFLOPs for a 0.7\% accuracy gain, while moving to WTConvNeXt-S adds only 4M parameters and 0.1 GFLOPs for a 0.5\% accuracy gain.

\subsection{Semantic Segmentation}\label{subsec:semseg}
We evaluate WTConvNeXt as a backbone for UperNet~\cite{Xiao:ECCV:2018:upernet} for the ADE20K~\cite{Zhou:IJCV:2019:ade20k} semantic segmentation task. We use MMSegmentation~\cite{Mmseg:2020} for UperNet's implementation, training, and evaluation. {The training follows the exact configuration for ConvNeXt without any parameter tuning.} We use the 80K and 160K iterations training scheme for the 120- and 300-epoch pre-trained models from \autoref{subsec:imagenet1k}, respectively, and report the mean intersection over union (mIoU) index with single-scale testing. 
{\autoref{tab:semseg} presents the results and shows an improvement of 0.3-0.6\% in mIoU when using WTConv.}

\subimport{./tables/}{segdet}

\subsection{Object Detection}\label{subsec:det}
We also evaluate WTConvNeXt as a backbone for Cascade Mask R-CNN~\cite{Cai:PAMI:2019:cascademaskrcnn} on the COCO dataset~\cite{Lin:ECCV:2014:coco}. We use MMDetection~\cite{Mmdet:2019} for Cascade Mask R-CNN's implementation, training, and evaluation. The training follows the exact configuration for ConvNeXt without any parameter tuning. We use the 1x and 3x fine-tuning schedules for the 120- and 300-epoch pre-trained models, respectively, and report box and mask average precision (AP). The results are presented in \autoref{tab:det_short}, where we see a considerable improvement, as AP$^{\mathrm{box}}$ and AP$^{\mathrm{mask}}$ both increase by 0.6-0.7\%. Detailed results are available at \textbf{Appendix F}.

\subsection{WTConv Analysis}\label{subsec:analysis}
\subsubsection{Scalability.}
We construct a small-scale scalability analysis for the task of classification on ImageNet-50/100/200~\cite{Van:ECCV:2020:scan,Ronen:CVPR:2022:DeepDPM}, which are subsets of ImageNet~\cite{Deng:CVPR:2009:imagenet} containing 50/100/200 classes, respectively.
In this experiment, we use MobilenetV2~\cite{Sandler:CVPR:2018:mobilenetv2} where each depth-wise convolution is replaced with RepLK~\cite{Ding:CVPR:2022:ScalingUpKernels}, GFNet \cite{Ra:NIPS2021:gfnet}, FFC \cite{chi:nips:2020:ffc}, and the proposed WTConv layer. We set WTConv's kernel sizes to $3\times3$.
For RepLK, we use the closest possible kernel size to WTConv's receptive field, \eg, $13\times13$ against 2-levels WTConv with a receptive field of $12\times12$.
GFNet and FFC are Fourier-based methods. GFnet's global filter layer requires $h \cdot w$ parameters per channel, where $(w,h)$ is the spatial dimension of the input, and is thus highly over-parameterized, especially on MobileNetV2 where the first few layers have an input of size $112^2$. In contrast, FFC uses the same weights across different frequencies, and therefore, it is not directly dependent on $(w,h)$ as GFNet is.
The training parameters are specified in \textbf{Appendix B}.

\subimport{./tables/}{scalability}

The results, summarized in \autoref{tab:scalability}, show that WTConv scales better than RepLK when increasing the receptive field. We assume this is due to insufficient data to support the large number of trainable parameters of the RepLK layers. This also aligns with the findings of~\cite{Liu:ICLR:2023:slak} for ImageNet-1K, where simply increasing the filter size in RepLK hurts the result. 
GFNet severely suffers from its over-parameterization, and its results drop significantly. FFC performs better, albeit the limited frequency mixing hurts its results.

\subsubsection{Robustness.}
We perform robustness evaluation for classification on ImageNet-C/$\overline{\mathrm{C}}$ \cite{Hendrycks:ICLR:2019:imagenetc, Mintun:NIPS:2021:imagenetcbar}, ImageNet-R \cite{Hendrycks:ICCV:2021:imagenetr}, ImageNet-A \cite{Hendrycks:CVPR:2021:imageneta}, and ImageNet-Sketch \cite{Wang:NIPS:2019:imagenetsk}. We report mean corruption error for ImageNet-C, corruption error for ImageNet-$\overline{\mathrm{C}}$, and top-1 accuracy for all other benchmarks. 
We also evaluate object detection on COCO under corruption~\cite{Michaelis:2019:detectionrobustness}, measured in mean and relative performance under corruption (mPC and rPC).
We report results for models trained using the 300-epoch schedule from \autoref{subsec:imagenet1k} without modifications or fine-tuning.

\subimport{./tables/}{robust_combined}

\autoref{tab:robustness} and \autoref{tab:robust_det} summarizes the results. Note that even though WTConvNeXt accuracy is $0.3-0.4\%$ above ConvNeXt in ImageNet-1K, the accuracy gain in most of the robustness datasets is above $1\%$ and gets as high as $2.2\%$. A similar trend is evident in corrupted object detection, which can be explained by the improved response to low frequencies~\cite{Li:PLOS:2023:robustlowfreq}.
More detailed tables and qualitative examples are provided in \textbf{Appendix G}.

\subsubsection{Shape-bias.}
We use the \emph{modelvshuman} benchmark~\cite{Geirhos:2021:humanvisionbenchmark} to quantify the improvement in shape bias (\ie, the fraction of predictions made based on shapes rather than on textures). Increased shape bias is associated with human perception and, therefore, considered more desirable. 

\subimport{./figures/}{analysis}

The results, presented in \autoref{fig:shapebias}, confirm our assumption that WTConv renders the network more shape-biased, increasing the fraction of ``shape" decisions by 8-12\%. Note that even the smaller WTConvNeXt-T responds better to shapes than the larger ConvNeXt networks, even though the latter score better at ImageNet-1k accuracy.
This is most likely due to the increased emphasis on the lower frequencies induced by WTConv, as shapes are generally associated with low frequencies, while textures are associated with high frequencies. 
The quantitative results are available in \textbf{Appendix E}

\subsubsection{Effective Receptive Field.}
We evaluate the contribution of WTConv to ConvNeXt-T's ERF~\cite{Luo:NIPS:2016:erf}, using the code provided by~\cite{Ding:CVPR:2022:ScalingUpKernels}. 
Theoretically, in CNNs, the ERF is proportional to $\Ocal(K\sqrt{L})$~\cite{Luo:NIPS:2016:erf}, where $K$ is the kernel size and $L$ is the depth of the network. 
However, since we introduce WT for an increased receptive field while using a smaller kernel, we assume it holds when considering $K$ as the size of the receptive field induced by the layer.
The empirical evaluation of the ERF includes sampling 50 random images from ImageNet's validation set resized to $1024 \times 1024$. Then, for each image, calculate each pixel's contribution, measured by the gradient, to the central point of the feature map produced by the last layer. 
The result is visualized in \autoref{fig:erf}, where high-contribution pixels are brighter. We note that WTConv has a nearly-global ERF despite having fewer parameters than RepLK and SLaK.

\subsubsection{Ablation Study.}
We conduct an ablation study to see how different configurations of the WTConv layer impact the final result.
We train WTConvNeXt-T for 120 epochs on ImageNet-1K as described in \autoref{subsec:imagenet1k}, with various configurations. 
First, we experiment with different combinations of WT levels and kernel sizes; note that ConvNeXt's convolutions operate on inputs of resolutions $56^2, 28^2, 14^2, 7^2$ (for $224^2$ input), allowing for maximal WT levels of $5,4,3,2$ respectively. Second, we evaluate the contribution of the high- and low-frequency components by using only one of the sets of high/low in the convolution each time. Lastly, we train the model with different wavelet bases. 

\Autoref{tab:ablation} shows the results of all the described configurations.
Here, increasing the levels and kernel size is mostly beneficial. We also see that using each frequency band separately improves the model's performance; however, using both is superior. 
The results confirm that the Haar WT is sufficient, although exploring other bases may improve performance. We leave this for future work.

\subimport{./tables/}{ablation}

%% file: tables/imagenet_convnext_120.tex
\begin{table}[ht]
    \centering
    \caption{Classification accuracy on ImageNet-1k using a 120-epoch training schedule.}
    \setlength{\tabcolsep}{4pt}
    \begin{tabular}{lccc}
         \toprule
         Configuration          & Receptive field & D-W Param. & Top-1  \\
         \midrule
         ConvNeXt-T             & [7,7,7,7]       & 0.32M      & 81.0     \\
         \quad+ VAN$^\dagger$ \cite{Guo:CVM:2023:VAN}          & [21,21,21,21]   & 0.38M      & 81.1     \\
         \quad+ GFNet$^\ddagger$ \cite{Ra:NIPS2021:gfnet}      & Global   & 4.29M      & 81.2     \\
         \quad+ RepLK \cite{Ding:CVPR:2022:ScalingUpKernels}   & [31,29,27,13]   & 3.84M      & 81.5     \\
         \quad+ SLaK$^\dagger$ \cite{Liu:ICLR:2023:slak} & [51,49,47,13]   & 2.52M      & 81.5     \\
         \quad+ WTConv         & [160,80,40,20]  & 2.04M      & \textbf{81.7}     \\
         \bottomrule
    \end{tabular}\\
    {\footnotesize
    $\dagger$ - using only kernel decomposition.\quad 
    $\ddagger$ - FFT-based method.
    }
    \label{tab:convnextimagenet_short}
\end{table}

%% file: tables/imagenet_convnext_300.tex
\begin{table}[ht]
    \centering
    \caption{Classification accuracy on ImageNet-1k using a 300-epoch training schedule.}    
    \setlength{\tabcolsep}{8pt}
    \begin{tabular}{lcccc}
         \toprule
         Configuration                               & Param. & FLOPs & Top-1 \\
         \midrule
         Swin-T \cite{Liu:ICCV:2021:swintransformer} & 28M   & 4.5G   & 81.3  \\
         ConvNeXt-T \cite{Liu:CVPR:2022:convnext}    & 29M   & 4.5G   & 82.1  \\
         WTConvNeXt-T                                & 30M   & 4.5G   & \textbf{82.5} \\
         \midrule
         Swin-S \cite{Liu:ICCV:2021:swintransformer} & 50M   & 8.7G   & 83.0  \\
         ConvNeXt-S \cite{Liu:CVPR:2022:convnext}    & 50M   & 8.7G   & 83.1  \\
         WTConvNeXt-S                                & 54M   & 8.8G   & \textbf{83.6} \\
         \midrule
         Swin-B \cite{Liu:ICCV:2021:swintransformer} & 88M   & 15.4G   & 83.5  \\
         ConvNeXt-B \cite{Liu:CVPR:2022:convnext}    & 89M   & 15.4G   & 83.8  \\
         WTConvNeXt-B                                & 93M   & 15.5G   & \textbf{84.1} \\
         \bottomrule
    \end{tabular}
    \label{tab:convnextimagenet_long}
\end{table}

%% file: tables/segdet.tex
\begin{table}[ht]
\centering
\begin{minipage}[t]{0.49\textwidth}\centering
    \setlength{\tabcolsep}{1pt}
    \caption{ADE20K validation results using UperNet \cite{Xiao:ECCV:2018:upernet}. FLOPs are based on an input size of $2048\times512$.}
    \begin{tabular}{lccc}
         \toprule
         Configuration                       & Param.     & FLOPs    & mIoU  \\
         \midrule
         \multicolumn{4}{c}{120-e pre-train + 80K finetune} \\
         \midrule
         ConvNeXt-T                          & 60M        & 939G     & 44.6     \\
         \quad+ RepLK     & 64M        & 975G     & 45.0     \\
         WTConvNeXt-T    & 62M        & 939G     & \textbf{45.4}  \\
         \midrule
         \multicolumn{4}{c}{300-e pre-train + 160K finetune} \\
         \midrule
         Swin-T                              & 60M        & 945G     & 44.5     \\
         ConvNeXt-T                          & 60M        & 939G     & 46.0     \\
         WTConvNeXt-T   & 62M        & 939G     & \textbf{46.6} \\
         \midrule
         Swin-S                              & 81M        & 1038G    & 47.6     \\
         ConvNeXt-S                          & 82M        & 1027G    & 48.7     \\
         WTConvNeXt-S    & 86M        & 1028G    & \textbf{49.0}     \\
         \bottomrule
    \end{tabular}
    \label{tab:semseg}
\end{minipage}
\hfill
\begin{minipage}[t]{0.49\textwidth}\centering
    \setlength{\tabcolsep}{1pt}
    \caption{COCO validation results using Cascade Mask-RCNN \cite{Cai:PAMI:2019:cascademaskrcnn}. FLOPs are based on an input size of $1280\times800$}
    \begin{tabular}{lccc}
         \toprule
         Configuration         & FLOPs & $\mathrm{AP}^{\mathrm{box}}$ & $\mathrm{AP}^{\mathrm{mask}}$  \\
         \midrule
         \multicolumn{4}{c}{120-e pre-train + 1x finetune} \\
         \midrule
         ConvNeXt-T            & 741G & 47.3 & 41.1 \\
         \quad+ RepLK    & 776G & 47.8 & 41.4 \\
         WTConvNeXt-T          & 741G & \textbf{47.9} & \textbf{41.5} \\
         \midrule
         \multicolumn{4}{c}{300-e pre-train + 3x finetune} \\
         \midrule
         Swin-T         & 745G & 50.4 & 43.7 \\
         ConvNeXt-T     & 741G & 50.4 & 43.7 \\
         WTConvNeXt-T   & 741G & \textbf{51.0} & \textbf{44.4} \\
         \midrule
         Swin-S         & 838G & 51.9 & 45.0 \\
         ConvNeXt-S     & 827G & 51.9 & 45.0 \\
         WTConvNeXt-S   & 827G & \textbf{52.6} & \textbf{45.6} \\
         \bottomrule
    \end{tabular}
    \label{tab:det_short}
\end{minipage}
\end{table}

%% file: tables/scalability.tex
\begin{table}[ht]
    \centering
    \setlength{\tabcolsep}{4.5pt}
    \caption{Classification accuracy on ImageNet-50/100/200, using MobileNetV2 with different depth-wise convolutions. Param./c is the parameter count per channel.}
    \begin{tabular}{lcccc}
        \toprule
        Conv Type & Param./c & IN-50 & IN-100 & IN-200 \\
        \midrule
         Baseline $3\times 3$          &   9 & 85.08 & 84.64 & 82.38 \\
        \midrule
         RepLK $7\times 7$   &  58 & 86.16 & 85.02 & 83.31 \\
         RepLK $13\times 13$ & 178 & 85.28 & 85.26 & 83.50 \\
         RepLK $25\times 25$ & 634 & 84.96 & 84.94 & 83.38 \\
         GFNet               &  $h \cdot w$ & 54.68 & 55.72 & 56.65 \\
         FFC                 &  14 & 84.92 & 84.64 & 82.96 \\
         \midrule
         WTConv 1 level      &  45 & \textbf{86.24} & 85.14 & 83.29 \\
         WTConv 2 levels     &  81 & 86.04 & 85.36 & 83.45 \\
         WTConv 3 levels     & 117 & 85.96 & \textbf{85.44} & \textbf{83.84} \\
         \bottomrule
    \end{tabular}
    \label{tab:scalability}
\end{table}

%% file: tables/robust_combined.tex
\begin{table}[ht]
\centering
\begin{minipage}[t]{0.55\textwidth}\centering
    \setlength{\tabcolsep}{2pt}
    \centering
    \caption{Robustness to corruption in classification over different benchmarks. ImageNet-$\mathrm{C}/\bar{\mathrm{C}}$ are measured in corruption error, and ImageNet-A/R/SK in top-1 accuracy.}
    \begin{tabular}{lccccc}
         \toprule
         Model           & C $\downarrow$ & $\overline{\mathrm{C}}$ $\downarrow$ & A $\uparrow$ & R $\uparrow$ & SK $\uparrow$ \\
         \midrule
         ConvNeXt-T      & 53.2 & 40.0 & 24.2 & 47.2 & 35.1 \\
         WTConvNeXt-T    & \textbf{52.0} & \textbf{38.0} & \textbf{25.3} & \textbf{48.0} & \textbf{35.8} \\
         \midrule
         ConvNeXt-B      & 46.8 & 34.4 & 36.7 & 51.3 & 39.4 \\
         WTConvNeXt-B    & \textbf{45.6} & \textbf{32.2} & \textbf{36.9} & \textbf{52.6} & \textbf{39.9} \\
         \bottomrule
    \end{tabular}
    \label{tab:robustness}
\end{minipage}
\hfill
\begin{minipage}[t]{0.43\textwidth}\centering
    \setlength{\tabcolsep}{3pt}
    \centering
    \caption{Robustness to corruptions in object detection measured in mean and relative performance under corruption (mPC and rPC).}
    \begin{tabular}{lccc}
         \toprule
         Model        & AP    & mPC   & rPC  \\
         \midrule
         ConvNeXt-T   & 50.4  & 31.8  & 63.2 \\
         WTConvNeXt-T & \textbf{51.0}  & \textbf{34.2}  & \textbf{67.1} \\
         \midrule
         ConvNeXt-S   & 51.9  & 35.2  & 67.8 \\
         WTConvNeXt-S & \textbf{52.6}  & \textbf{36.9}  & \textbf{70.3} \\
         \bottomrule
    \end{tabular}
    \label{tab:robust_det}
\end{minipage}
\end{table}

%% file: figures/analysis.tex
\begin{figure}[t]
\centering
    \includegraphics[width=0.6\linewidth]{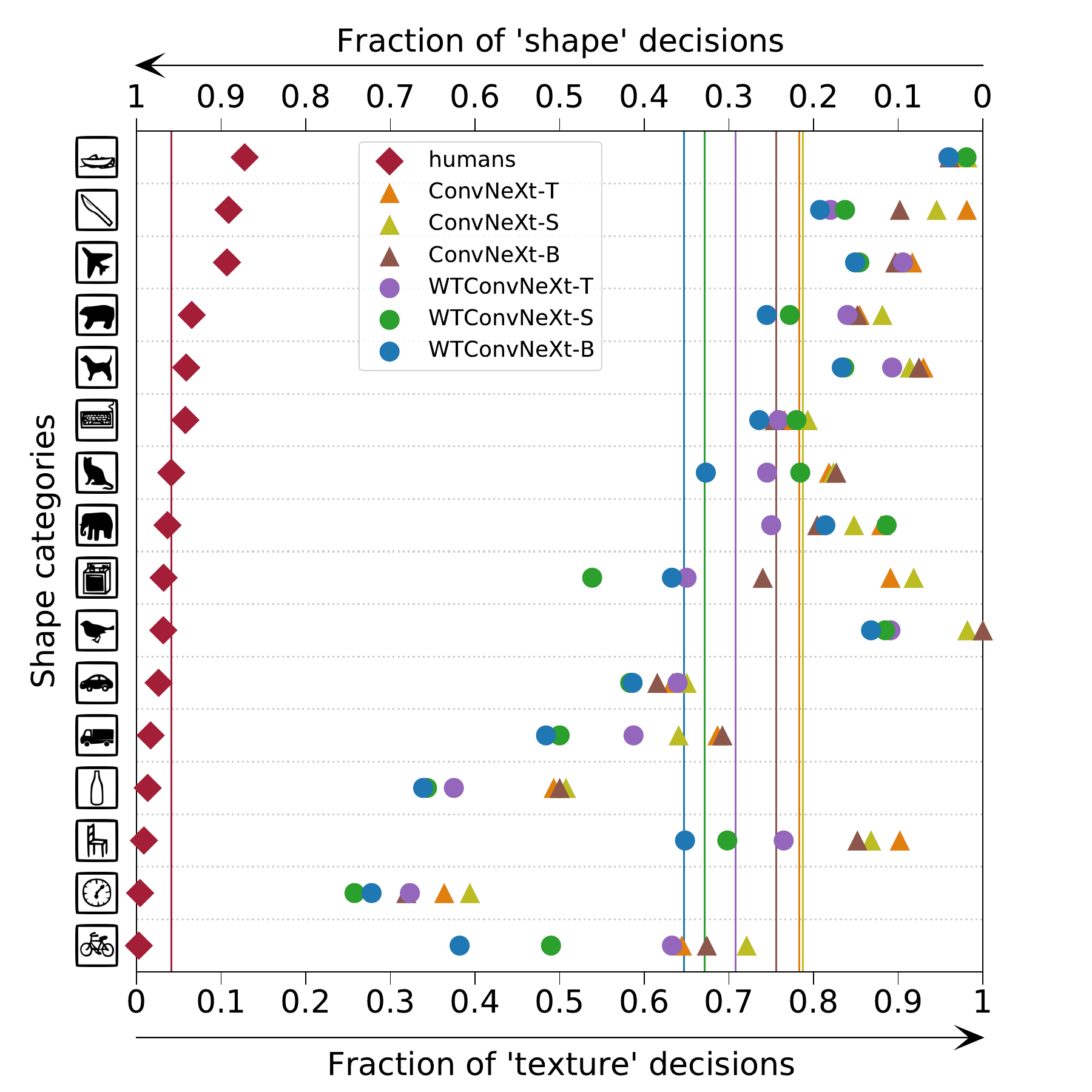}
\caption{Shape bias comparison of ConvNeXt-T/S/B and WTConvNeXt-T/S/B over 16 categories. The vertical line is the average across categories.}
\label{fig:shapebias}
\end{figure}

%% file: tables/ablation.tex
\begin{table}[ht]
    \centering
    \setlength{\tabcolsep}{3pt}
    \caption{Ablation study with WTConvNeXt-T. Comparing different configurations of WTConv.}
    \begin{tabular}{cccccc}
         \toprule
         Levels       & Kernel size & Wavelet   & Receptive field & D-W Param. & Top-1  \\
         \midrule
         $[4,3,2,1]$  & $3\times3$  & Haar & [48,24,12,6] &  0.50M     & 81.24  \\
         $[4,3,2,1]$  & $5\times5$  & Haar & [80,40,20,10] &  1.38M     & 81.32  \\
         $[4,3,2,1]$  & $7\times7$  & Haar & [112,56,28,14] &  2.70M     & 81.56  \\
         $[5,4,3,2]$  & $3\times3$  & Haar & [96,48,24,12] &  0.73M     & 81.49   \\
         $[5,4,3,2]$  & $5\times5$  & Haar & [160,80,40,20] &  2.04M     & \textbf{81.75}  \\
         $[5,4,3,2]$  & $7\times7$  & Haar & [224,112,28,14] &  3.99M     & 81.69  \\
         \midrule
         $[5,4,3,2]$  & $5\times5$ Lows  & Haar & [160,80,40,20] &  0.63M     &  81.46 \\
         $[5,4,3,2]$  & $5\times5$ Highs & Haar & [160,80,40,20] &  1.57M     &  81.24 \\
         \midrule
         $[5,4,3,2]$  & $5\times5$  & db2       & [320,160,80,40] &  2.04M     &  81.68 \\
         $[5,4,3,2]$  & $5\times5$  & db3       & [640,320,160,80] &  2.04M     &  81.56 \\
         \bottomrule
    \end{tabular}
    \label{tab:ablation}
\end{table}

%% file: conclusion.tex
\section{Limitations}
Although the WTConv layer does not require many FLOPs, its running time can be relatively high within existing frameworks. This is due to the overhead for the multiple sequential operations (WT-conv-IWT), which may be more costly than the calculation itself. We note, however, that this can be alleviated by using a specialized implementation, \eg, performing WT in parallel to convolution in each level to reduce memory reads or performing WT and IWT in-place to reduce memory allocations. More implementation details are provided in \textbf{Appendix C}.

\section{Conclusion}
In this work, we utilize wavelet transforms to introduce WTConv, a drop-in replacement for depth-wise convolutions that  achieves a larger receptive field and better captures low frequencies in the input. 
Using WTConv, one can configure a spatial mixing of a global receptive field in a pure convolutional way.
We demonstrate empirically that WTConv substantially increases the CNNs' effective receptive fields, improve the CNNs' shape bias, renders the networks more robust to corruptions, and yields better performance for various vision tasks.

%% file: appendix.tex
\section{The Algorithm} \label{apnd:alg}
\begin{algorithm}
\caption{WTConv}
\begin{algorithmic}
    \STATE {\bfseries Input:} $X\in\mathbb{R}^{C \times N_w\times N_h}$
    \STATE $X_{LL}^{(0)} = X$
    \STATE $Y_{LL}^{(0)} = \mathrm{Conv}(W^{(0)}, X_{LL}^{(0)})$
    \FOR{$i=1,\dots,\ell$}
    \STATE $X_{LL}^{(i)},X_{LH}^{(i)},X_{HL}^{(i)},X_{HH}^{(i)} = \mathrm{WT}(X_{LL}^{(i)})$
    \STATE $Y_{LL}^{(i)},Y_{LH}^{(i)},Y_{HL}^{(i)},Y_{HH}^{(i)} = \mathrm{Conv}(W^{(i)},(X_{LL}^{(i)},X_{LH}^{(i)},X_{HL}^{(i)},X_{HH}^{(i)}))$
    \ENDFOR
    \STATE $Z^{(\ell+1)}=0$
    \FOR{$i=\ell,\dots,1$}
    \STATE $Z^{(i)}=\mathrm{IWT}(Y_{LL}^{(i)}+Z^{(i+1)},Y_{LH}^{(i)},Y_{HL}^{(i)},Y_{HH}^{(i)})$
    \ENDFOR
    \STATE $Z^{(0)} = Y_{LL}^{(0)} + Z^{(1)}$
    \RETURN $Z^{(0)}$
\end{algorithmic}
\end{algorithm}

\section{Training Parameters}\label{apnd:trainingparams}
\subsection{Imagenet-50/100/200}\label{apnd:imnetsubset_params}
The training process is similar to~\cite{Ding:CVPR:2022:ScalingUpKernels}, using a 2-GPU setup, an SGD optimizer with a momentum of 0.9, a batch size of 32 per GPU, input resolution 224$\times$224, weight decay of $4\cdot 10^{-5}$, and a 5-epoch linear warmup followed by cosine annealing for 100 epochs. The initial learning rate was adjusted to 0.025, taking the number of GPUs into consideration. The implementation is based on~\cite{PytorchClassRecipe}.

\subsection{ImageNet-1K}\label{apnd:imnet1k_params}
We follow~\cite{Liu:CVPR:2022:convnext,Ding:CVPR:2022:ScalingUpKernels} for the 300-epochs training schedule with AdamW~\cite{Loshchilov:2017:adamw} with momentum 0.9 and weight decay 0.05, batch size 4096, learning rate $4\cdot 10^{-3}$, a 20-epoch linear warm-up followed by cosine annealing, RandAugment~\cite{Cubuk:CVPR:2020:randaugment}, label smoothing~\cite{Szegedy:CVPR:2016:inception} coefficient 0.1, mixup~\cite{Zhang:2017:mixup} with $\alpha=0.8$, CutMix~\cite{Yun:ICCV:2019:cutmix} with $\alpha=1.0$, Random Erasing~\cite{Zhong:AAAI:2020:randerase} with probability 25\%, Stochastic Depth~\cite{Huang:ECCV:2016:stochasticdepth} with a drop-path rate of 10\%/40\%/50\% for T/S/B variations respectively, and 
an exponential moving average (EMA) with a decay factor of 0.9999. We also provide comparisons for a shorter training schedule 
(which follows~\cite{Ding:CVPR:2022:ScalingUpKernels}) whose configuration is similar to the one above, except that it uses  120 epochs, batch size 2048, 10-epoch warm-up, and no EMA.

\section{Naive implementation running times}
The implementation used for all our experiments is fairly naive, and many improvements can be made with it. For example, in addition to what we described in Section 5, the Haar wavelet is currently implemented in our model as a regular convolution, which includes multiplications by 1 and -1 as FP32. These operations, however, can be changed to summation and substractions and can be done for all the levels at the same time for more efficient memory reads.

Using the naive implementation, we timed the throughput of ConvNeXt and WTConvNeXt using a single RTX3090 for 300 batches of size 64 after a GPU warmup of 50 batches. \Autoref{tab:naive_times} presents the throughput, measured in images per second. As can be seen, even with the most naive and unoptimized version of the layer, WTConvNeXt has a throughput of 66-70\% of the original network.

\begin{table}[ht]
    \centering
    \caption{Naive implementation throughput measured by images per second.}
    \setlength{\tabcolsep}{10pt}
    \begin{tabular}{lc}
         \toprule
         Model        & Images/sec \\
         \midrule
         ConvNeXt-T   & 976.65 \\
         WTConvNeXt-T & 652.76 \\
         \midrule
         ConvNeXt-S   & 565.52 \\
         WTConvNeXt-S & 384.91 \\
         \midrule
         ConvNeXt-B   & 371.99 \\
         WTConvNeXt-B & 260.12 \\
         \bottomrule
    \end{tabular}
    \label{tab:naive_times}
\end{table}

\section{ImageNet-1K - additional results}
To demonstrate WTConv compatibility, we incorporate it within two additional networks, GhostNet \cite{Han:CVPR:2020:ghostnet} and EfficientNet \cite{Tan:ICML:2019:efficientnet}. The number of WTConv levels is set to have a global receptive field at each stage \wrt input size of $224\times224$. The training procedure is as described in \autoref{apnd:imnet1k_params} for the 120-epochs training schedule. The results are presented in \autoref{tab:ghosteff}.

\section{quantitative shape-bias results}
The quantitative shape-bias results are reported in \autoref{tab:shapefraction}.

\begin{table}[ht]
\begin{minipage}[t]{0.49\textwidth}\centering
    \centering
    \caption{Additional ImageNet-1K results with 120-epochs schedule.}
    \begin{tabular}{lcc}
        \toprule
        Configuration  & Params. & Top-1 \\
        \midrule
        GhostNet       & 5.2M & 67.41 \\
        \quad +WTConv  & 5.7M & \textbf{68.22} \\
        \midrule
        EfficientNet-B0  & 5.3M & 73.25 \\
        \quad +WTConv    & 6.4M & \textbf{74.34} \\
        \bottomrule
    \end{tabular}
    \label{tab:ghosteff}
\end{minipage}
\hfill
\begin{minipage}[t]{0.49\textwidth}\centering
    \centering
    \caption{The average fraction of ``shape" decisions for every network.}
    \begin{tabular}{lccc}
    \toprule
                 & T    & S    & B    \\
    \midrule
    ConvNeXt     & 21.7\% & 21.2\% & 24.5\% \\
    WTConvNeXt   & \textbf{29.2\%} & \textbf{32.9\%} & \textbf{35.3\%} \\
    \bottomrule
    \end{tabular}
    \label{tab:shapefraction}
\end{minipage}
\end{table}

\section{Object detection - additional results}
\Autoref{tab:det} provides the detailed results for COCO object detection experiment, as described in \textbf{\S 4.3}.

\subimport{./tables/}{detection}

\clearpage

\section{Robustness - additional results}
\Autoref{tab:imnet_c} and \autoref{tab:imnet_c_bar} provide the detailed results for ImageNet-C and ImageNet-$\overline{\mathrm{C}}$, respectively. \Autoref{Fig:road_corrupt_det}, \Autoref{Fig:zebras_corrupt_det}, \Autoref{Fig:bikers_corrupt_det}, and \Autoref{Fig:boat_corrupt_det} are qualitative examples of object detection under different types of corruption. These examples show that as the corruption is more severe, WTConvNeXt loses fewer details.

\begin{table}[ht]
    \centering
    \setlength{\tabcolsep}{2.3pt}
    \caption{ImageNet-C detailed results.}
    \begin{tabular}{lcc|ccc|cccc}
         \toprule
         \multicolumn{3}{c|}{} & \multicolumn{3}{c|}{Noise} & \multicolumn{4}{c}{Blur} \\
         Model           & \footnotesize Err & \footnotesize mCE & \footnotesize Gauss. & \footnotesize Shot & \footnotesize Impulse & \footnotesize Defocus & \footnotesize Glass & \footnotesize Motion & \footnotesize Zoom\\
         \midrule
         ConvNeXt-T      & 41.6 & 53.2 & 37.1 & 38.7 & 39.6 & \textbf{50.4} & 63.0 & 43.6 & 54.2 \\
         WTConvNeXt-T    & \textbf{40.7} & \textbf{52.0} & \textbf{36.6} & \textbf{37.6} & \textbf{37.6} & 50.9 & \textbf{60.5} & \textbf{43.4} & \textbf{52.0} \\
         \midrule
         ConvNeXt-B      & 36.6 & 46.8 & 32.0 & 33.1 & 31.7 & 46.0 & 56.3 & 38.4 & 47.2 \\
         WTConvNeXt-B    & \textbf{35.6} & \textbf{45.6} & \textbf{30.0} & \textbf{31.3} & \textbf{30.9} & \textbf{45.7} & \textbf{55.8} & \textbf{36.3} & \textbf{44.6} \\
         \bottomrule
    \end{tabular}
    \vspace{1em}
    \begin{tabular}{l|cccc|cccc}
         \toprule
          & \multicolumn{4}{c|}{Weather} & \multicolumn{4}{c}{Digital} \\
         Model           & \footnotesize Snow & \footnotesize Frost & \footnotesize Fog & \footnotesize Bright & \footnotesize Contrast & \footnotesize Elastic & \footnotesize Pixel & \footnotesize JPEG\\
         \midrule
         ConvNeXt-T      & 42.1 & 38.5 & \textbf{38.4} & \textbf{23.3} & \textbf{32.1} & 46.0 & 44.5 & \textbf{32.6} \\
         WTConvNeXt-T    & \textbf{41.4} & \textbf{37.9} & 39.1 & \textbf{23.3} & 32.4 & \textbf{45.8} & \textbf{40.0} & \textbf{32.6} \\
         \midrule
         ConvNeXt-B      & 38.0 & \textbf{33.3} & 35.0 & 21.1 & \textbf{28.1} & 41.0 & 37.0 & 30.2 \\
         WTConvNeXt-B    & \textbf{35.4} & 33.8 & \textbf{32.6} & \textbf{21.0} & 31.8 & \textbf{39.4} & \textbf{36.3} & \textbf{29.4} \\
         \bottomrule
    \end{tabular}
    \label{tab:imnet_c}
\end{table}

\begin{table}[ht]
    \centering
    \setlength{\tabcolsep}{3.5pt}
    \caption{ImageNet-$\overline{\mathrm{C}}$ detailed results.}
    \begin{tabular}{lc|cccccccccc}
         \toprule
         Model           & Err & \small BSmpl & \small Plsm & \small Ckbd & \small CSin & \small SFrq & \small Brwn & \small Prln & \small Sprk & \small ISprk & \small Rfrac \\
         \midrule
         ConvNeXt-T      & 40.0 & 30.8 & 43.8 & 33.0 & 69.0 & 53.5 & 24.9 & \textbf{25.3} & 26.8 & 59.4 & \textbf{33.9} \\
         WTConvNeXt-T    & \textbf{38.0} & \textbf{29.8} & \textbf{40.5} & \textbf{31.3} & \textbf{60.0} & \textbf{49.3} & \textbf{24.8} & 25.5 & \textbf{26.3} & \textbf{59.0} & 34.0 \\
         \midrule
         ConvNeXt-B      & 34.4 & 25.6 & 40.6 & 29.2 & 56.6 & 38.6 & 22.5 & 22.7 & 24.5 & 53.3 & 30.7 \\
         WTConvNeXt-B    & \textbf{32.2} & \textbf{25.0} & \textbf{36.0} & \textbf{27.3} & \textbf{46.3} & \textbf{38.4} & \textbf{22.2} & \textbf{22.6} & \textbf{24.0} & \textbf{50.1} & \textbf{29.5} \\
         \bottomrule
    \end{tabular}
    \label{tab:imnet_c_bar}
\end{table}

\begin{figure}[ht]
    \centering
    \setlength\tabcolsep{1.5pt}
    \begin{tabular}{ccc}
    
    \includegraphics[width=.33\textwidth]{figures/imgs/road_2_motion_blur_2.jpg} &
    \includegraphics[width=.33\textwidth]{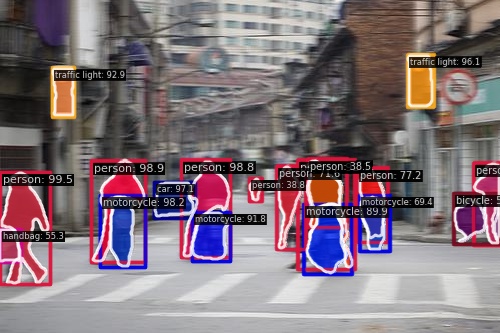} &
    \includegraphics[width=.33\textwidth]{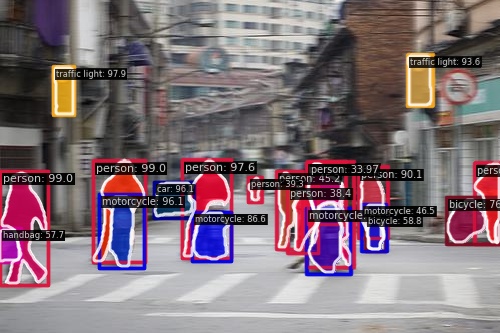}
    \\
    \includegraphics[width=.33\textwidth]{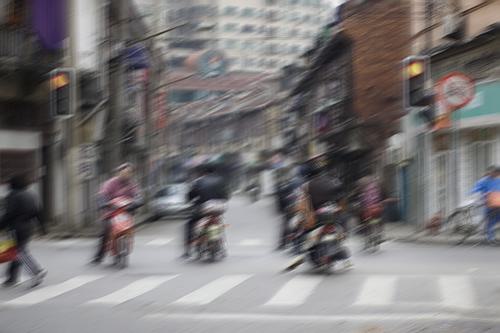} &
    \includegraphics[width=.33\textwidth]{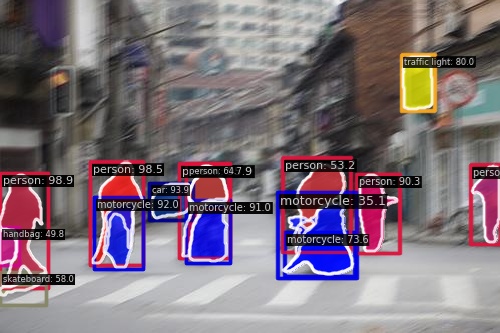} &
    \includegraphics[width=.33\textwidth]{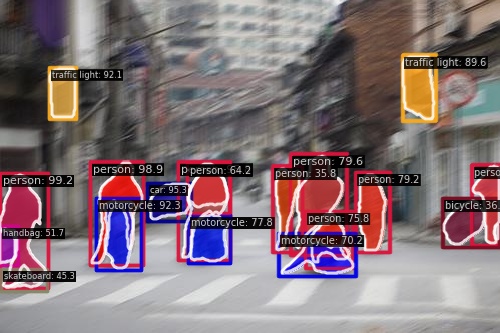}
    \\
    \includegraphics[width=.33\textwidth]{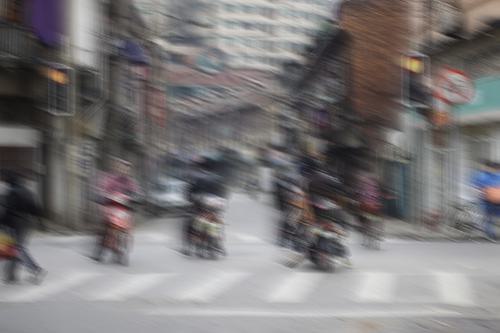} &
    \includegraphics[width=.33\textwidth]{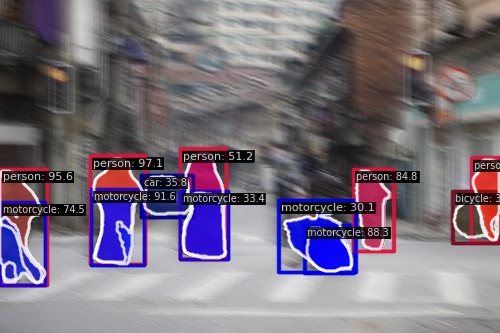} &
    \includegraphics[width=.33\textwidth]{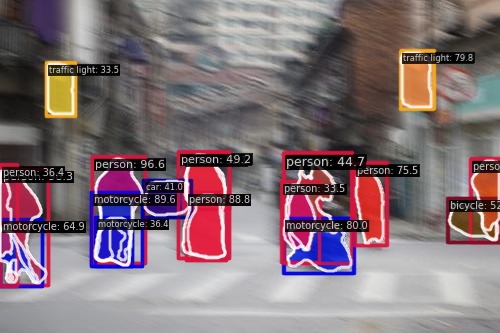}\\
    Input image & ConvNeXt-S backbone & WTConvNeXt-S backbone
    \end{tabular}
    \caption{Motion blur corruption example. Each row represents an increased corruption severity. Left to right: the corrupted input, ConvNeXt-S backbone detection, WTConvNeXt-S backbone detection. Note that WTConvNeXt detects the traffic light even at the worst corruption level.}
    \label{Fig:road_corrupt_det}
\end{figure}

\begin{figure}[ht]
    \centering
    \setlength\tabcolsep{1.5pt}
    \begin{tabular}{ccc}
    
    \includegraphics[width=.33\textwidth]{figures/imgs/zebras_1_pixelate_2.jpg} &
    \includegraphics[width=.33\textwidth]{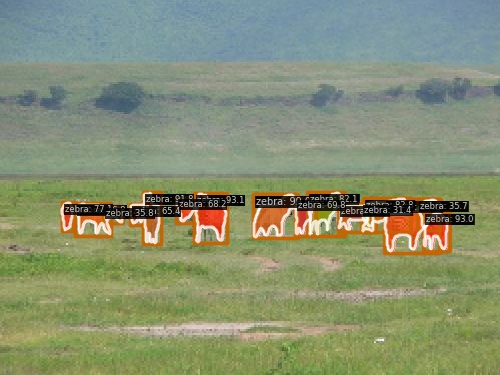} &
    \includegraphics[width=.33\textwidth]{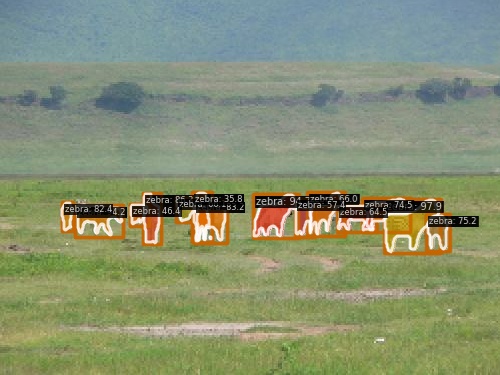}
    \\
    \includegraphics[width=.33\textwidth]{figures/imgs/zebras_1_pixelate_3.jpg} &
    \includegraphics[width=.33\textwidth]{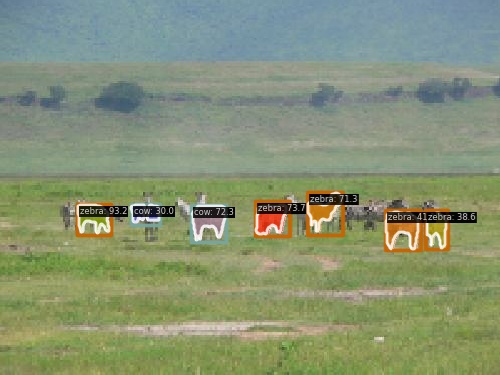} &
    \includegraphics[width=.33\textwidth]{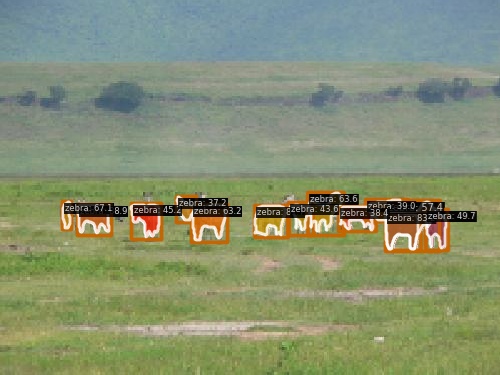}
    \\
    \includegraphics[width=.33\textwidth]{figures/imgs/zebras_1_pixelate_4.jpg} &
    \includegraphics[width=.33\textwidth]{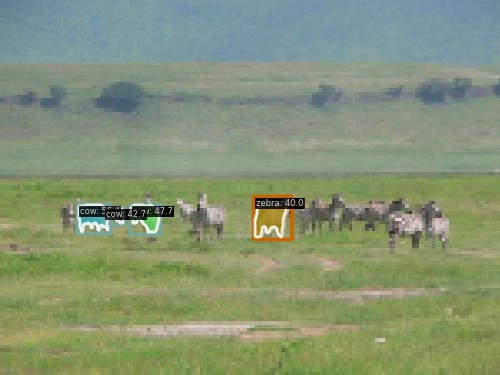} &
    \includegraphics[width=.33\textwidth]{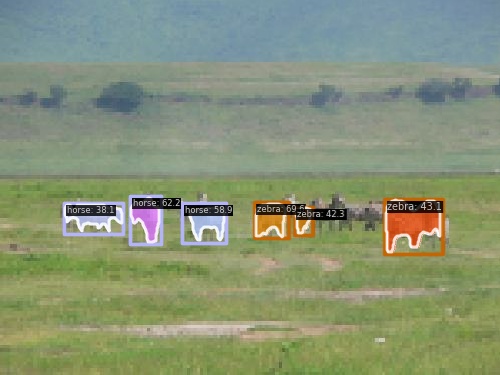}\\
    Input image & ConvNeXt-S backbone & WTConvNeXt-S backbone
    \end{tabular}
    \caption{Pixelate corruption example. Each row represents an increased corruption severity. Left to right: the corrupted input, ConvNeXt-S backbone detection, WTConvNeXt-S backbone detection. Note that WTConvNeXt detects more zebras than ConvNeXt and classifies them more accurately.}
    \label{Fig:zebras_corrupt_det}
\end{figure}

\begin{figure}[ht]
    \centering
    \setlength\tabcolsep{1.5pt}
    \begin{tabular}{ccc}
    
    \includegraphics[width=.33\textwidth]{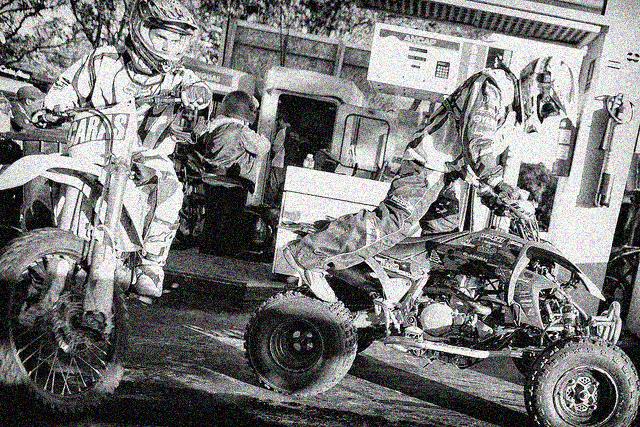} &
    \includegraphics[width=.33\textwidth]{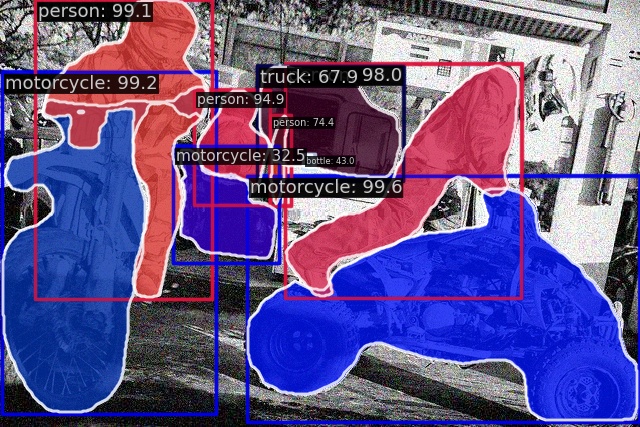} &
    \includegraphics[width=.33\textwidth]{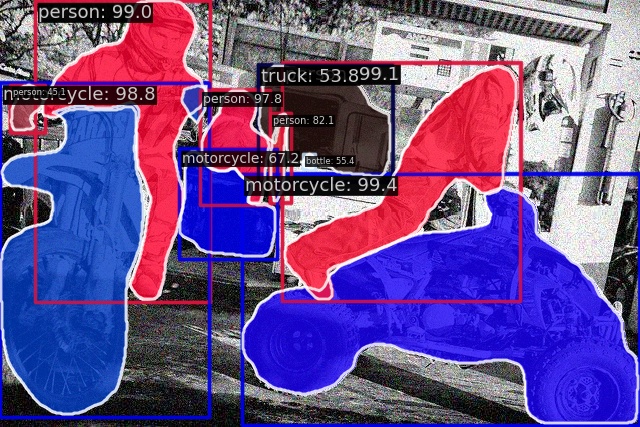}
    \\
    \includegraphics[width=.33\textwidth]{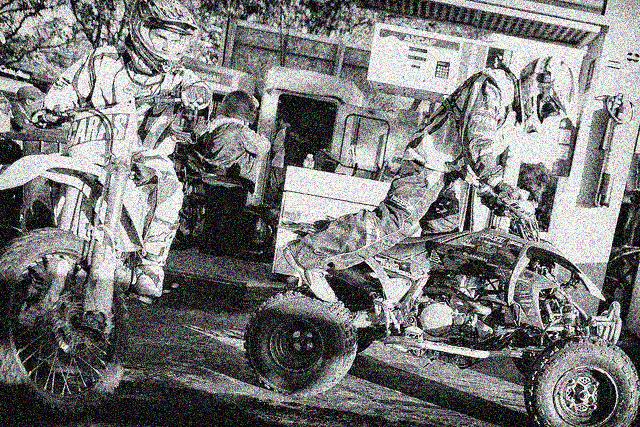} &
    \includegraphics[width=.33\textwidth]{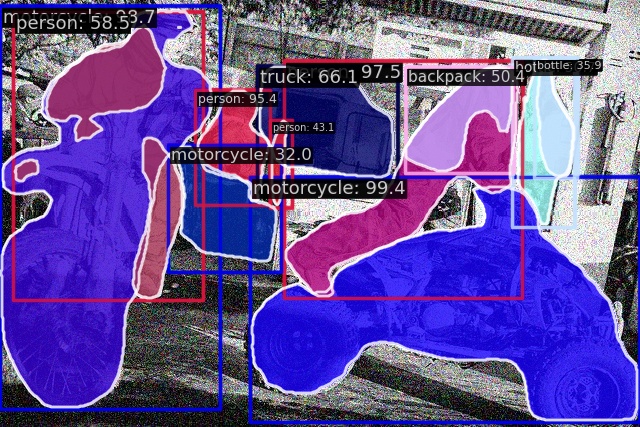} &
    \includegraphics[width=.33\textwidth]{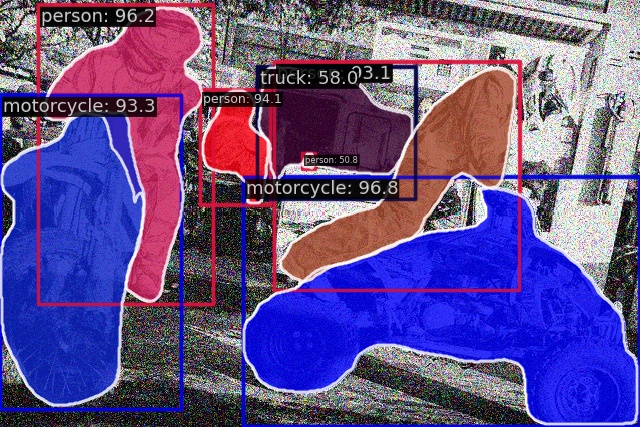}
    \\
    \includegraphics[width=.33\textwidth]{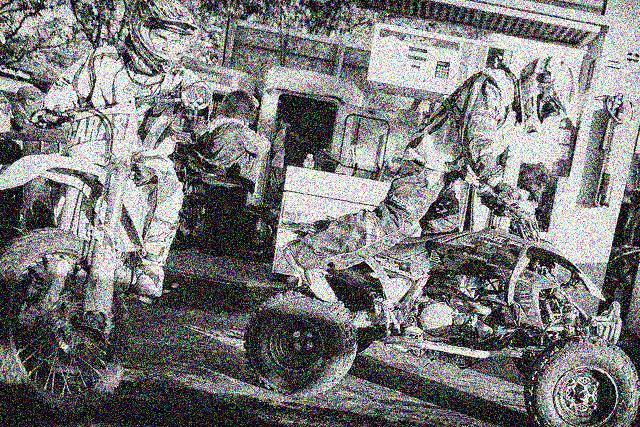} &
    \includegraphics[width=.33\textwidth]{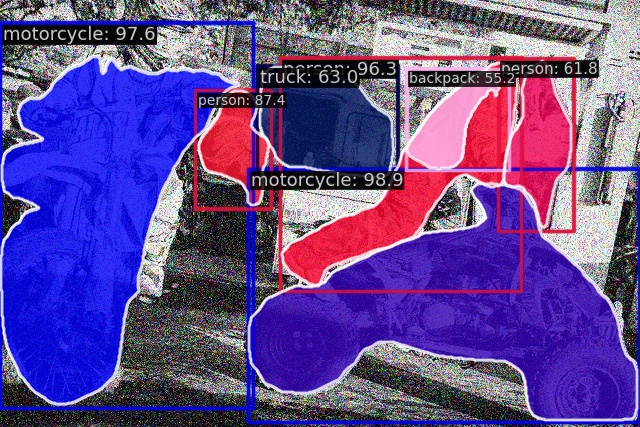} &
    \includegraphics[width=.33\textwidth]{figures/imgs/bikers_gauss_4_wt.jpg}\\
    Input image & ConvNeXt-S backbone & WTConvNeXt-S backbone
    \end{tabular}
    \caption{Gaussian noise corruption example. Each row represents an increased corruption severity. Left to right: the corrupted input, ConvNeXt-S backbone detection, WTConvNeXt-S backbone detection. WTConvNeXt clearly captures the object better as the corruption level increases.}
    \label{Fig:bikers_corrupt_det}
\end{figure}

\begin{figure}[ht]
    \centering
    \setlength\tabcolsep{1.5pt}
    \begin{tabular}{ccc}
    
    \includegraphics[width=.33\textwidth]{figures/imgs/boat_glass_blur_2.jpg} &
    \includegraphics[width=.33\textwidth]{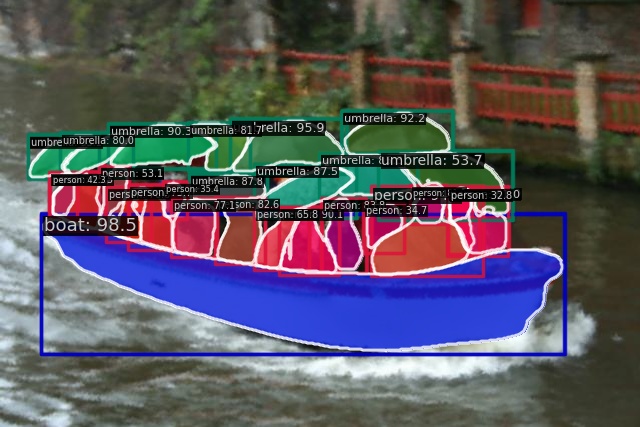} &
    \includegraphics[width=.33\textwidth]{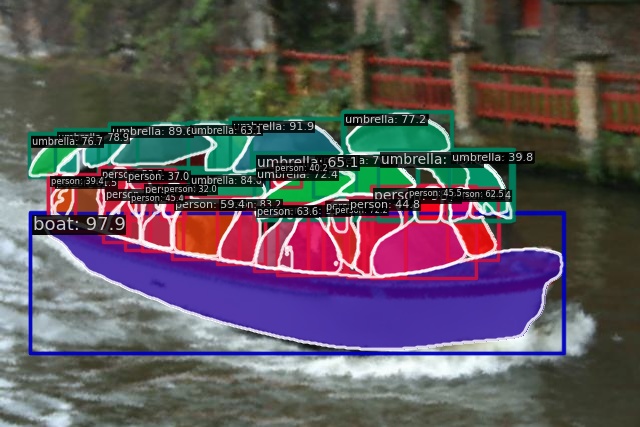}
    \\
    \includegraphics[width=.33\textwidth]{figures/imgs/boat_glass_blur_3.jpg} &
    \includegraphics[width=.33\textwidth]{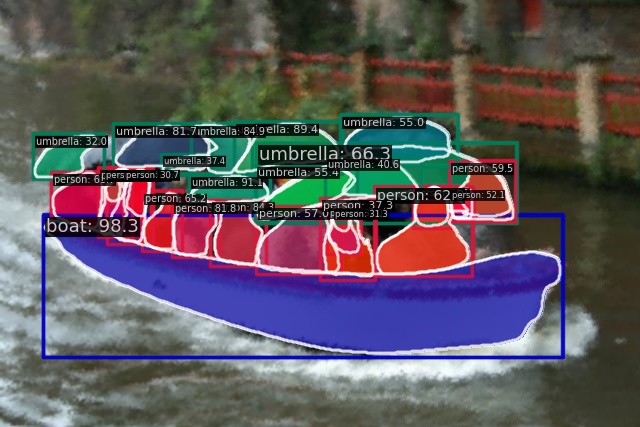} &
    \includegraphics[width=.33\textwidth]{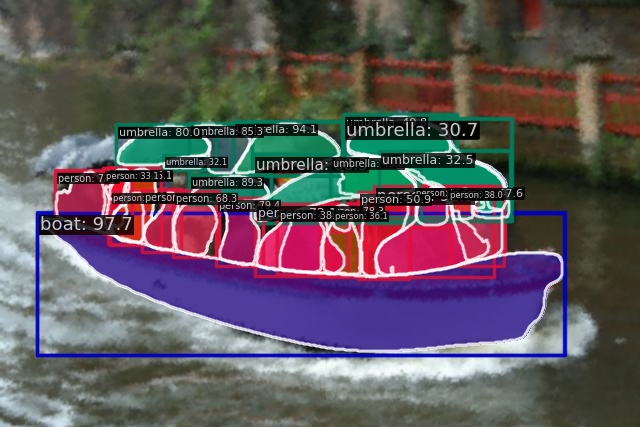}
    \\
    \includegraphics[width=.33\textwidth]{figures/imgs/boat_glass_blur_4.jpg} &
    \includegraphics[width=.33\textwidth]{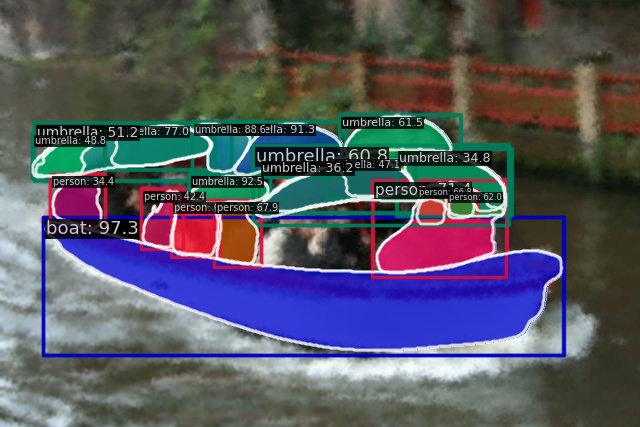} &
    \includegraphics[width=.33\textwidth]{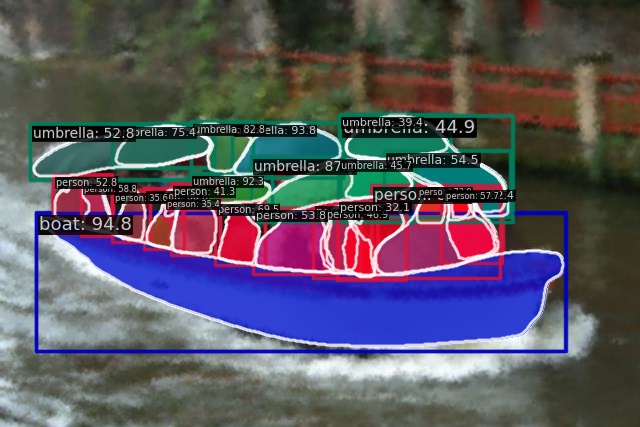}\\
    Input image & ConvNeXt-S backbone & WTConvNeXt-S backbone
    \end{tabular}
    \caption{Glass blur corruption example. Each row represents an increased corruption severity. Left to right: the corrupted input, ConvNeXt-S backbone detection, WTConvNeXt-S backbone detection. Note that ConvNeXt misses some of the persons on the boat, while WTConvNeXt classifies them correctly.}
    \label{Fig:boat_corrupt_det}
\end{figure}

%% file: tables/detection.tex
\begin{table*}[ht]
    \centering
    \setlength{\tabcolsep}{4pt}
    \caption{COCO object detection and segmentation results using Cascade Mask-RCNN \cite{Cai:PAMI:2019:cascademaskrcnn} with different backbones pre-trained on ImageNet-1k using 120/300-epochs training schedule, and fine-tuned with 1x/3x schedule.}
    \begin{tabular}{lcccccc}
         \toprule
         Configuration                       & $\mathrm{AP}^{\mathrm{box}}$ & $\mathrm{AP}^{\mathrm{box}}_{50}$ & $\mathrm{AP}^{\mathrm{box}}_{75}$ & $\mathrm{AP}^{\mathrm{mask}}$ & $\mathrm{AP}^{\mathrm{mask}}_{50}$ & $\mathrm{AP}^{\mathrm{mask}}_{75}$  \\
         \midrule
         \multicolumn{7}{c}{120-epoch pre-train + 1x finetune} \\
         \midrule
         ConvNeXt-T            & 47.3 & 65.9 & 51.5 & 41.1 & 63.2 & 44.4 \\
         ConvNeXt-T + RepLK    & 47.8 & 66.7 & 52.0 & 41.4 & 63.9 & 44.7 \\
         WTConvNeXt-T          & \textbf{47.9} & 66.7 & 52.0 & \textbf{41.5} & 64.0 & 44.9 \\
         \midrule
         \multicolumn{7}{c}{300-epoch pre-train + 3x finetune} \\
         \midrule
         Swin-T         & 50.4 & 69.2 & 54.7 & 43.7 & 66.6 & 47.3 \\
         ConvNeXt-T     & 50.4 & 69.1 & 54.8 & 43.7 & 66.5 & 47.3 \\
         WTConvNeXt-T   & \textbf{51.0} & 69.9 & 55.3 & \textbf{44.4} & 67.2 & 48.1 \\
         \midrule
         Swin-S         & 51.9 & 70.7 & 56.5 & 45.0 & 68.2 & 48.8 \\
         ConvNeXt-S     & 51.9 & 70.8 & 56.4 & 45.0 & 68.4 & 49.1 \\
         WTConvNeXt-S   & \textbf{52.6} & 71.5 & 57.1 & \textbf{45.6} & 69.0 & 49.5 \\
         \bottomrule
    \end{tabular}
    \label{tab:det}
\end{table*}